\documentclass[runningheads]{llncs}
\usepackage[T1]{fontenc}
\usepackage{amsmath,amsfonts}
\usepackage{algorithmic}
\usepackage{array}
\usepackage{textcomp}
\usepackage{stfloats}
\usepackage{url}
\usepackage{verbatim}
\usepackage{graphicx}
\usepackage{cite}
\usepackage{microtype}

\usepackage[dvipsnames]{xcolor}

\usepackage{subcaption}
\usepackage{graphicx}
\usepackage{comment}
\usepackage{amsmath}
\usepackage{amssymb}
\usepackage{amsfonts}
\usepackage{pifont}
\usepackage{booktabs}
\usepackage{algorithmic}
\usepackage{textcomp}
\usepackage{cite}
\usepackage{xcolor,soul}
\usepackage[normalem]{ulem}
\usepackage{kotex}
\usepackage{multirow, array}
\usepackage{url}
\usepackage{footnote}
\makesavenoteenv{tabular}
\usepackage{relsize}
\usepackage[linesnumberedhidden,ruled,vlined]{algorithm2e}
\usepackage{adjustbox}
\usepackage{caption}
\usepackage{stfloats}
\usepackage[accsupp]{axessibility}
\usepackage[symbol]{footmisc}

\let\subparagraph\relax 
\usepackage[compact]{titlesec}
\titlespacing*{\section}{0pt}{1.2ex}{0.6ex}
\titlespacing*{\subsection}{0pt}{1.0ex}{0.5ex}
\usepackage[compact]{titlesec}
\usepackage{microtype}
\usepackage[font=scriptsize, labelfont=bf]{caption}
\usepackage{enumitem}
\setlist[itemize]{noitemsep, topsep=0pt}
\usepackage{graphicx}
\begin{document}
\title{Optimizing Three Critical Factors for Practical and Effective OOD Detection Fine-tuning}
%
%
\author{Hyunjun Choi\inst{1,2}\orcidID{0000-0002-1413-1285}\and
JaeHo Chung\inst{1}\orcidID{0009-0008-2332-117X} \and
Hawook Jeong\inst{2}\orcidID{0009-0003-5935-7055}} 

%
\authorrunning{H. Choi et al.} 
\titlerunning{Optimizing Three Factors for Practical OOD Detection Fine-tuning} 
%
\institute{ ASRI, ECE., Seoul National University, South Korea 
\email{\{numb7315,jaehochung\}@snu.ac.kr}
\and
RideFlux Inc., South Korea \email{hawook@rideflux.com}}

\maketitle              

\begin{abstract}

In out-of-distribution (OOD) detection, fine-tuning with auxiliary outlier data often improves detection performance at the cost of classification accuracy. This trade-off stems from the loss of the original in-distribution (ID) distribution during fine-tuning. To establish a more practical and effective paradigm, we optimize three critical factors: model reminder, data sampling, and representation learning. We propose: (1) Self-Knowledge Distillation (SKD) to mitigate accuracy reduction; (2) Semi-hard Outlier Sampling (SOS) to improve detection efficiency with minimal data; and (3) Outlier-aware Supervised Contrastive Learning (OSCL) to promote ID-OOD separability. Optimizing these factors produces cumulative gains, boosting both OOD detection performance and classification accuracy. Our framework outperforms existing methods across diverse benchmarks, particularly in long-tailed scenarios, providing a robust baseline for real-world OOD detection.    

\end{abstract}

\section{Introduction}

Assigning high confidence to out-of-distribution (OOD) data is a crucial problem in safety-critical fields such as autonomous driving~\cite{kendall2017uncertainties} and medicine~\cite{leibig2017leveraging}; to address this, a benchmark and the maximum softmax probability (MSP) baseline were proposed~\cite{hendrycks2016baseline}.
Many efforts have been made to improve OOD detection performance such as using a good observer~\cite{lee2018simple,liu2020energy,sun2021react,wang2022vim, sun2022out} and utilizing an auxiliary outlier data~\cite{hendrycks2018deep,papadopoulos2021outlier,liu2020energy,choi2023}.
In our work, we pay attention to the approach of using auxiliary outlier data to improve performance on OOD detection through fine-tuning. The first paper to employ auxiliary data is Outlier Exposure (OE)~\cite{hendrycks2018deep}. OE leverages cross-entropy loss for the primary training data and regularization loss for the auxiliary data. The regularization loss in OE is derived from the cross-entropy loss incurred by assigning a uniform label to the auxiliary data. After that, various types of regularization loss have been proposed subsequently such as OECC~\cite{papadopoulos2021outlier} with calibration loss,  EnergyOE~\cite{liu2020energy} with energy score and double hinge regularization loss.
Recently, Balanced EnergyOE~\cite{choi2023} has been proposed to address the imbalance problem in outlier data.
However, these methods mainly focus on designing regularization losses to improve OOD detection, often leading to decreased classification accuracy compared to the base model~\cite{yang2021generalized}. Notably, in non-ideal datasets like the long-tailed CIFAR10, OE exhibits significant accuracy drops, falling to 69.73\% from a baseline of 73.28\%. Empirical analysis indicates this stems from the catastrophic loss of the original ID distribution during fine-tuning (see Section~\ref{analysis_accuracy}).

To establish a more practical and effective fine-tuning paradigm, we optimize three critical factors: model reminder, data sampling, and representation learning.
First, we propose Self-Knowledge Distillation (SKD) as a "model reminder" to prevent the catastrophic loss of original in-distribution (ID) knowledge that typically occurs during outlier-driven fine-tuning. During the fine-tuning process, the parameters of the teacher model are frozen, and the student model updates its weights by referencing fixed targets derived from the teacher through Softening Targets distillation loss~\cite{hinton2015distilling}. By "reminding" the model of its original ID distribution, SKD effectively preserves the ID negative energy gap and offsets accuracy reduction, leading to cumulative gains in both classification performance and OOD detection (see Section~\ref{analysis_OOD}).

Second, we optimize the quantity and quality of outlier training data to improve efficiency. We observe that utilizing only 10\% (30K images) of the TinyImages-300K dataset~\cite{hendrycks2018deep} is sufficient for effective OOD detection, matching the performance of PASCL~\cite{wang2022partial}. Regarding quality, we identify that "hardness"—defined as the maximum softmax probability (MSP) of outliers on a pre-trained model—is a critical performance driver.
While training on high-hardness samples improves OOD detection, excessive hardness eventually diminishes both accuracy and detection performance. To navigate this trade-off, we propose Semi-hard Outlier Sampling (SOS), which selects outliers within an optimal quantile range to achieve a superior balance between detection power and classification accuracy.


    

\begin{figure}[t!]
\centering
    \begin{subfigure}[b]{0.48\linewidth}
        \centering
        \includegraphics[width=\linewidth]{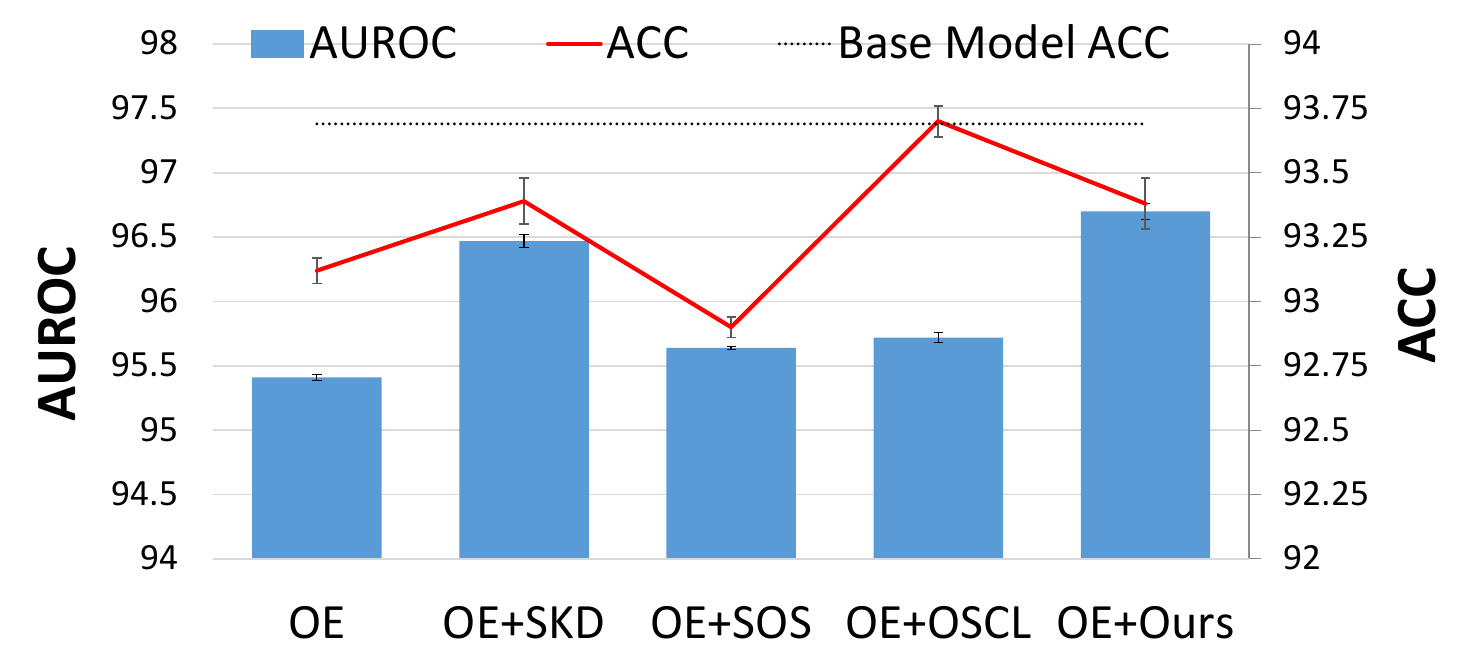}
        \caption{CIFAR10}
        \label{intro_10}
    \end{subfigure}
    \hfill 
    \begin{subfigure}[b]{0.48\linewidth}
        \centering
        \includegraphics[width=\linewidth]{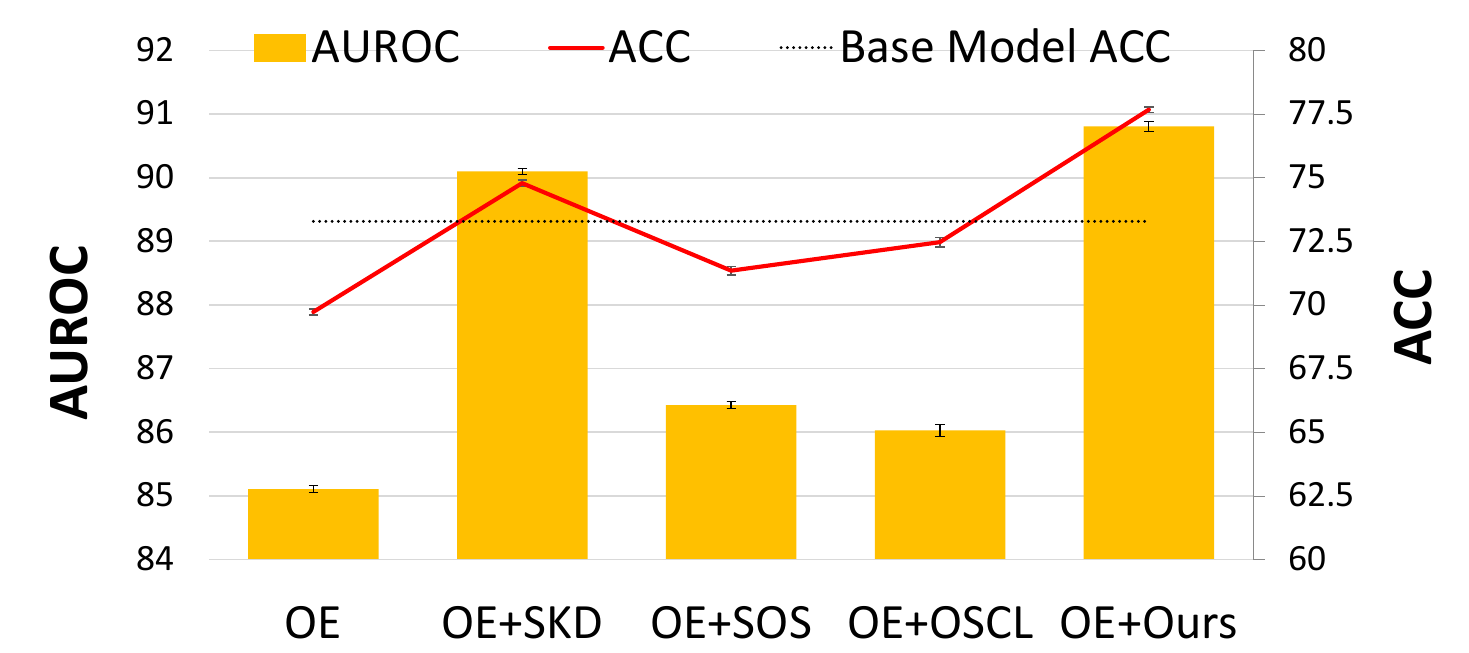}
        \caption{LT-CIFAR10}
        \label{intro_10_lt}
    \end{subfigure}
    
    \caption{\textbf{Effects of each of the three methods and our combined method.} Combining the three methods with the baseline OE results in improvements in both OOD detection performance (AUROC) and accuracy (ACC).}
    \label{intro_fig}
\end{figure}
Third, we enhance representation quality through Outlier-aware Supervised Contrastive Learning (OSCL). The key distinction of OSCL from traditional SCL~\cite{graf2021dissecting} is the explicit use of outlier data as negative samples to drive contrastive embeddings away from in-distribution samples. This promotes superior ID-OOD separability, as validated by our analysis using the CIDER~\cite{ming2022cider} metric in Section~\ref{analysis_OSCL}. Furthermore, we introduce a multi-batch transform to significantly boost the effectiveness of contrastive learning. As a result, the learned embeddings achieve inter-class dispersion with large angular distances among class prototypes while simultaneously promoting intra-class compactness.

Integrating these three factors produces cumulative improvements in both OOD detection and classification accuracy (Figure~\ref{intro_fig}). Our framework generalizes across ideal and non-ideal datasets; specifically, in long-tailed scenarios, it mitigates accuracy degradation and often surpasses the original base model's performance. These simple yet effective methods are orthogonal to existing fine-tuning frameworks and require no extra data or modules. Extensive validation on SC-OOD~\cite{yang2021semantically}, MOOD~\cite{lin2021mood}, and SYN~\cite{hendrycks2018deep} benchmarks—including integration with OE, OECC, and EnergyOE—confirms the wide applicability and potency of our approach in practical OOD detection.
\section{Related Works}

\noindent\textbf{OOD Detection using Auxiliary Data.} Outlier Exposure (OE)~\cite{hendrycks2018deep} established the paradigm of using non-overlapping auxiliary data as outliers, regularizing the model to produce uniform distributions on such samples. Subsequent works have refined this via diverse regularization losses, including confidence control in OECC~\cite{papadopoulos2021outlier}, energy-based hinge loss in EnergyOE~\cite{liu2020energy}, and adaptive losses for imbalanced outliers in Balanced EnergyOE~\cite{choi2023}. Our approach is orthogonal to these loss-centric methods, enhancing both OOD detection and classification accuracy across various fine-tuning frameworks.

\noindent\textbf{OOD Detection with Observer Mode.} Observer-mode methods derive OOD scores from pre-trained models without modifying the architecture or loss. Representative measures include MSP~\cite{hendrycks2016baseline}, Mahalanobis distance~\cite{lee2018simple}, and Free Energy~\cite{liu2020energy}, with recent enhancements like rectified activations (ReAct)~\cite{sun2021react} and nearest-neighbor search (kNN)~\cite{sun2022out}. While simple, these methods often exhibit dependencies on specific datasets or networks. In contrast, our fine-tuning approach provides more robust and reliable detection performance.

\noindent\textbf{OOD Detection with Contrastive Learning.} Contrastive learning has evolved from self-supervised frameworks like SimCLR~\cite{chen2020simple} to supervised settings (SCL)~\cite{khosla2020supervised} to improve representation quality. For long-tailed or complex distributions, methods such as BCL~\cite{zhu2022balanced}, CIDER~\cite{ming2022cider}, and PASCL~\cite{wang2022partial} have introduced balanced or asymmetric repulsion losses. Our proposed OSCL extends this by explicitly incorporating outliers as negative samples, driving ID-OOD separability further than traditional SCL-based approaches.
\section{Exploring Three Factors for Enhancing OOD Detection}

\subsection{Problem Statements}

 We can formalize a classifier as $\mathbf{z}(\mathbf{x}):\mathbb{R}^0D\to\mathbb{R}^K$
    , which maps an input image vector $\mathbf{x}$ with $D$ dimension to a real-valued vector (logit) with $K$ dimension where $K$ is the number of classes. $\mathbf{z}(\mathbf{x})=[z_{1}(\mathbf{x}),\dots,z_{K}(\mathbf{x})]$ and $\mathbf{y}$  is $K$-dimensional one-hot target vector. Probability vector $\mathbf{p}(\mathbf{x})=[p_{1}(\mathbf{x}),\dots,p_{K}(\mathbf{x})]$ is computed as $Softmax(\mathbf{z}(\mathbf{x}))$, which satisfies $\mathbf{1}^{T}\mathbf{p}(\mathbf{x})=1$ and $\mathbf{p}(\mathbf{x}) \geq 0$.    
    On the other hand, $\mathbf{z}(\mathbf{x}):\mathbb{R}^D\to\mathbb{R}^K$  is a composition of encoder $\overline{\mathbf{x}}=\mathbf{g}(\mathbf{x}):\mathbb{R}^D\to\mathbb{R}^L$  and linear classifier $\mathbf{w}(\overline{\mathbf{x}}):\mathbb{R}^L\to\mathbb{R}^K$, 
    i.e., $\mathbf{z}(\mathbf{x})=(\mathbf{w} \circ \mathbf{g})(\mathbf{x})$. 
Final projection layer $\mathbf{f}(\mathbf{x}):\mathbb{R}^D\to\mathbb{R}^N$ for contrastive embedding is defined as $\mathbf{f}(\mathbf{x})=(\mathbf{h} \circ \mathbf{g})(\mathbf{x})$, where we use another linear layer 
$\mathbf{h}(\overline{\mathbf{x}}):\mathbb{R}^L\to\mathbb{R}^K$ 
in composition of the feature $\overline{\mathbf{x}}=\mathbf{g}(\mathbf{x})$. 
Finally, $\mathbf{f}(\mathbf{x})=[f_{1}(\mathbf{x}),\dots,f_{N}(\mathbf{x})]$ and $L_{k}$-normalized contrastive feature is defined as $\widetilde{\mathbf{f}}(\mathbf{x})=[\frac{f_{1}(\mathbf{x})}{\lVert f \rVert_{k}},\dots,\frac{f_{N}(\mathbf{x})}{\lVert f \rVert_{k}}]^T$. The contrastive embedding $\widetilde{\mathbf{f}}$ lies on the unit hypersphere.

The model using outlier data for fine-tuning is trained by a loss function that combines the classification loss $L_{cls}$ for in-distribution data and the regularization loss $L_{reg}$ for the outlier data. The loss is written as 
    \begin{equation}
    \begin{split}
    \small
    \label{equation:general_OE}
    L&= L_{cls}+\lambda_{reg}L_{reg}\\
    &=\mathbb{E}_{(\mathbf{x},\mathbf{y})\sim{D_{in}^{train}}}[H(\mathbf{y},\mathbf{p}(\mathbf{x}))]+\lambda_{reg}L_{reg},
    \end{split}
    \end{equation} 
where, $H(\mathbf{a},\mathbf{b})$ is cross-entropy between probability distribution $\mathbf{a}$ and $\mathbf{b}$. $L_{reg}$ is designed depending on the Outlier Exposure (OE) strategy. In OE~\cite{hendrycks2018deep}, the regularization loss is used to encourage outlier data to have a uniform probability distribution $\mathbf{u}$.
\begin{equation} 
\label{equation:OE}
\small
L_{reg}=\mathbb{E}_{\mathbf{x}\sim{D_{out}^{train}}}[H(\mathbf{u},\mathbf{p}(\mathbf{x}))]. 
\end{equation}
The above fine-tuning model sacrifices classification accuracy in exchange for improved performance of OOD detection~\cite{yang2021generalized}.
In our work, to address this issue, we explore two additional regularization terms: Self-Knowledge Distillation loss and Outlier-aware Supervised Contrastive loss, along with a Semi-hard Outlier Sampling strategy. 

\subsection{Self-Knowledge Distillation}
\label{SKD}
To prevent the loss of in-distribution (ID) knowledge during fine-tuning, we propose \textbf{Self-Knowledge Distillation (SKD)} as a "model reminder". Our approach is primarily motivated by the analysis in Section~\ref{analysis_accuracy}, which reveals that standard OE fine-tuning tends to suppress the original ID distribution, leading to a significant degradation in classification accuracy. Unlike standard KD, our SKD operates within the same model by leveraging the original pre-trained parameters as a frozen teacher.
The student model updates its parameters by referencing fixed targets from the teacher model using \textbf{Softening Targets} distillation loss~\cite{hinton2015distilling}:
\begin{equation} 
    \small
    \label{equation:distill_before}
    \widetilde{p_{i}}(\mathbf{x};T_{KD})=\frac{exp(z_{i}(\mathbf{x})/T_{KD})}{\sum_{j}exp(z_{j}(\mathbf{x})/T_{KD})},
\end{equation}

\begin{equation} 
    \small
    \label{equation:distill}
    L_{KD}=T_{KD}^{2}H(\widetilde{\mathbf{p}}^{T}(\mathbf{x};T_{KD}),\widetilde{\mathbf{p}}^{S}(\mathbf{x};T_{KD})).
\end{equation}
$T_{KD}$ represents the scaling temperature, while $\widetilde{p_{i}}$ is scaled probability of the $i$-th component. 
$\widetilde{\mathbf{p}}^{T}$ and $\widetilde{\mathbf{p}}^{S}$ respectively denote the distribution of the teacher model and the student model.
The empirical effectiveness of incorporating SKD into the fine-tuning framework is evaluated in Section~\ref{combined}, where we observe consistent improvements in both OOD detection and classification accuracy. A detailed analysis regarding the energy gap preservation achieved by SKD is provided in Section~\ref{analysis_OOD}.
\subsection{Semi-hard Outlier Sampling}
\label{ODS}
\begin{figure}[t!]
\centering

\begin{subfigure}[b]{0.7\linewidth}
        \caption{}
        \includegraphics[width=\linewidth]{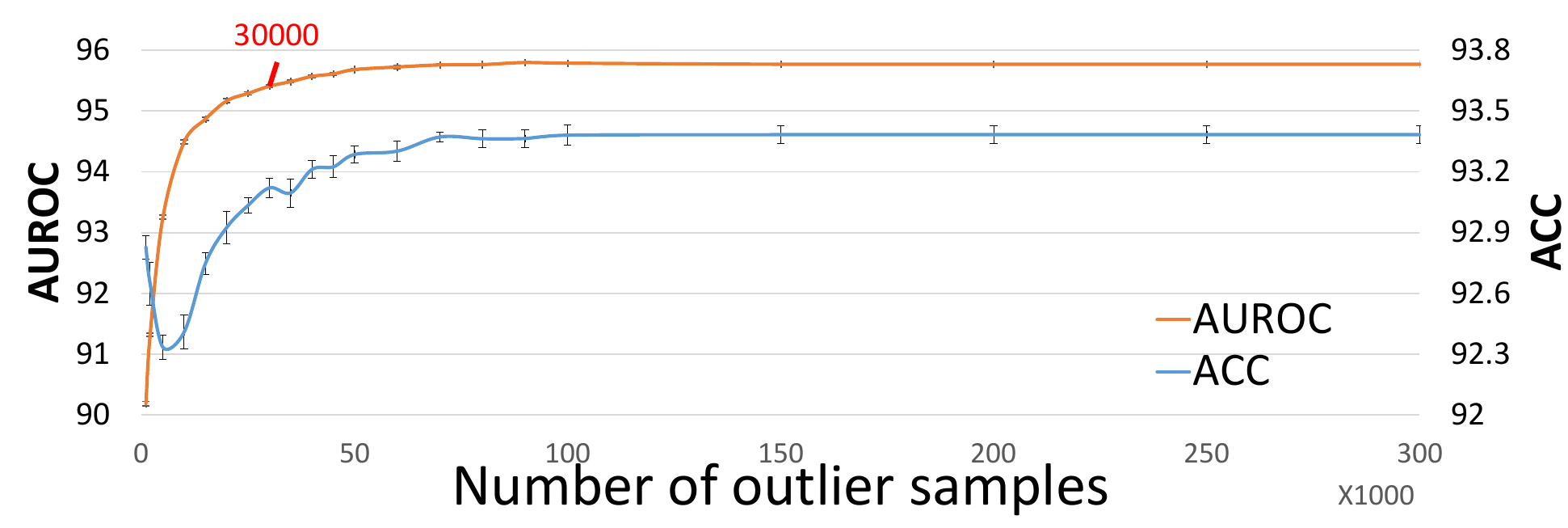}
        \label{fig_num_sample}
\end{subfigure}
\begin{subfigure}[b]{0.7\linewidth}
        \caption{}
        \includegraphics[width=\linewidth]{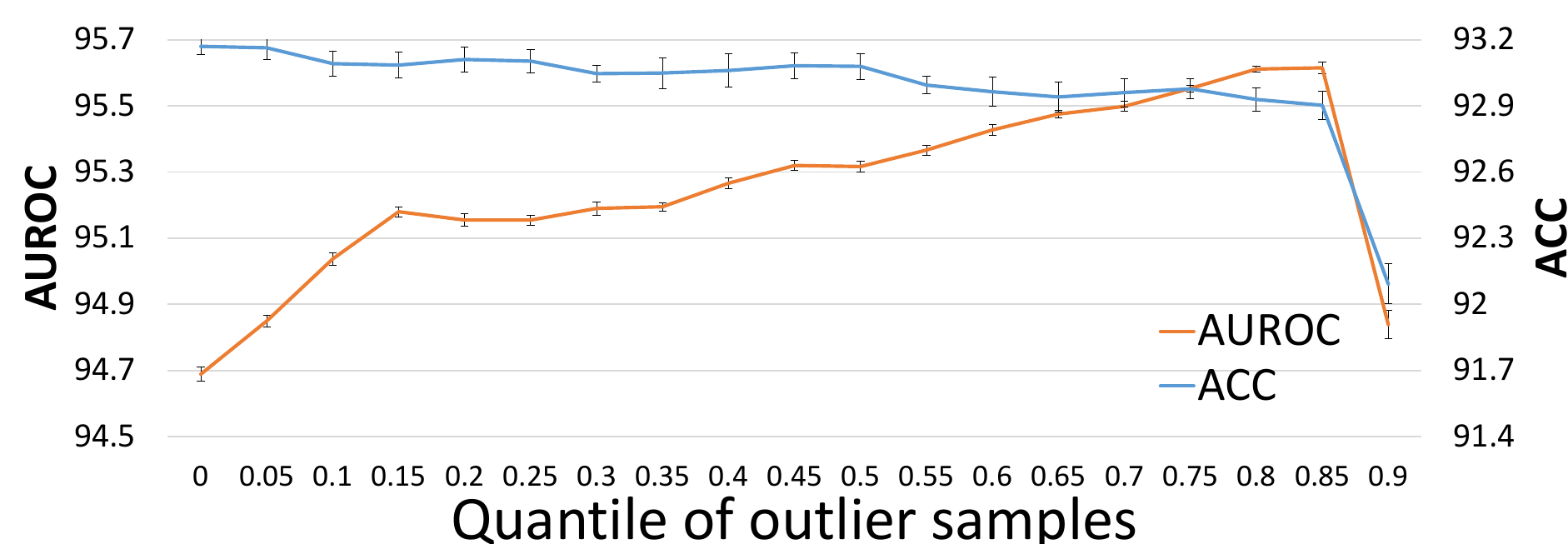}
        \label{fig_quantile}
\end{subfigure}
\caption{\textbf{Impact of outlier training samples' quantity (a) and quality (hardness) (b)} Figure \ref{fig_num_sample} shows the relation between outlier samples' quantity and performance for OOD detection (SC-OOD average AUROC) and classification (ACC). Figure \ref{fig_quantile} depicts the relation between outlier samples' hardness quantile and performance for a step size of 1.
}
\label{fig_num_quantile}
\end{figure}

Empirically we examine how outlier training data affects OOD detection performance in relation to classification accuracy from two different perspectives.
One perspective is the number of outlier data used in training, and the other is the level of difficulty of outlier data, referred to as hardness~\cite{bo2021hardness}.
As shown in Figure~\ref{fig_num_sample}, we observe that using only 10\% (30K images) of the TinyImages-300K~\cite{hendrycks2018deep} dataset as outlier data in OE can effectively achieve good OOD performance, which is the same setting as PASCL~\cite{wang2022partial}. 

Next, we investigate whether there is an optimal selection of 30K outlier images that achieve the best OOD detection performance. We discover the presence of optimal outlier data within a specific range of hardness. We define hardness as the maximum softmax probability (MSP) obtained by inferring outlier data on a pre-trained model. Hardness is defined as follows:
\begin{equation} 
    \small
    \label{equation:distill_before}
    hardness = \max\limits_{i\in \{1,2,\dots,K\} }p_{i}(\mathbf{x}).
\end{equation}

\begin{algorithm}[t]
\scriptsize
\SetKwInput{KwInput}{Input}
\SetKwInput{KwData}{Data}
\SetKwInput{KwOutput}{Output}
\DontPrintSemicolon 

\KwInput{$\mathbf{z}$: model, $m$: size, $q$: quantile, $step$: step}
\KwData{$D_{in}$: in-dist, $D_{out}^{all}$: outliers ($M$)}
\KwOutput{$D_{out}^{sampled}$ with size $m$}

\BlankLine
\textbf{Step 1: Sampling based on hardness} \;
Load weight of $\mathbf{z}$ and inference on $D_{out}^{all}$ \;
$S \leftarrow \{ \max_{i} p_{i}(\mathbf{x}) \mid \mathbf{x} \in D_{out}^{all} \}$ \;
$I \leftarrow \text{argsort}(S)$ \;
$D_{out}^{sampled} \leftarrow D_{out}^{all} [I[q \cdot M : q \cdot M + m : step]]$ \;

\BlankLine
\textbf{Step 2: Fine-tuning on sampled data} \;
\For{$t=1$ \KwTo $T$}{
    Load mini-batches $D_{mini,i} \subset D_{in}$ and $D_{mini,o} \subset D_{out}^{sampled}$ \;
    Update $\mathbf{z}$ by minimizing Eq.~(\ref{equation:general_OE}) \;
} 

\caption{Outlier Data Sampling}
\label{algo}
\end{algorithm}
Figure~\ref{fig_quantile} shows that while increasing outlier hardness generally improves OOD detection, it degrades classification accuracy, with a sharp performance drop at the highest quantiles. To navigate this trade-off, we propose \textbf{Semi-hard Outlier Sampling (SOS)}, which selects outliers within an optimal quantile range $q$.
As detailed in Algorithm~\ref{algo}, SOS consists of sampling outliers based on hardness and subsequent fine-tuning. We control the sampling difficulty through quantile $q$ and step-size $step$, selecting values that yield optimal OOD performance. We compare SOS against baseline OE using both fixed random samples (following PASCL~\cite{wang2022partial}) and random samples per run, with results averaged over 8 runs. The synergy between SOS and other factors is further analyzed in Section~\ref{combined}.

\subsection{Outlier-aware Supervised Contrastive Learning}
\label{SCL}

Naive integration of existing contrastive methods often fails to yield significant OOD improvements as shown in Table~\ref{C_comparison_result}. We propose \textbf{Outlier-aware Supervised Contrastive Learning (OSCL)} to improve representation quality by incorporating outlier data as negative samples. Unlike traditional SCL~\cite{khosla2020supervised}, OSCL explicitly drives contrastive embeddings away from in-distribution samples using outlier data. The loss is formulated as:
\begin{equation} 
    \small
    \label{equation:sc_total_OOD}
   L_{SC}= \sum_{i \in B_{in}}L_{i}, 
\end{equation}
\begin{equation} 
    \small
    \label{equation:sc_each_OOD}
   L_{i}= -\frac{\mathbf{1}_{\{|B^{in}_{y_{i}}|>1\}}}{|B^{in}_{y_{i}}|-1} \sum_{p \in B^{in}_{y_{i}} \setminus \{i\}}{\log \frac{\exp(\widetilde{\mathbf{f}}_{i}\cdot \widetilde{\mathbf{f}}_{p} / \tau_{sc}) }{ \sum_{k \in  B^{all} \setminus \{i\}}{ \exp (\widetilde{\mathbf{f}}_{i}\cdot \widetilde{\mathbf{f}}_{k} / \tau_{sc})}}},
\end{equation}
where the total batch $B^{all}$ is partitioned into ID ($B^{in}$) and outlier ($B^{out}$) sets, and $B^{in}_{y_i}$ contains ID samples matching label $y_i$. The total fine-tuning objective is defined as $L = L_{cls} + \lambda_{reg}L_{reg} + \lambda_{SC}L_{SC}$.

To further boost contrastive learning, we introduce a \textbf{multi-batch transform}, generalizing the two-batch approach of SimCLR~\cite{chen2020simple}. For each input $\mathbf{x}$, we apply $n$ transformations $t_1, \dots, t_n \sim \mathcal{T}$ to generate augmented samples $\widetilde{\mathbf{x}}_{k}=t_{k}(\mathbf{x})$, which are concatenated to form $\mathbf{x}_{multi-batch}(n)=[\widetilde{\mathbf{x}}_{1}, \dots, \widetilde{\mathbf{x}}_{n} ]$. This simple yet effective extension allows the model to leverage multiple augmentation views simultaneously for enhanced representation. In our experiments, we select the optimal $n \in \{2,4,6,8\}$. Detailed analyses of the loss, multi-batch synergy, and SCL comparisons are provided in Sections~\ref{combined}, \ref{analysis_OSCL}, and \ref{C_discussion}.

\begin{table}[t!]
\centering
\caption{Incremental Ablation Study of the three proposed factors on SC-OOD benchmark. We report Mean and std over 8 random runs. (SKD: Self-Knowledge Distillation, SOS: Semi-hard Outlier Sampling, OSCL: Outlier-aware Supervised Contrastive Learning)}
\label{tab:unified_ablation}
\setlength{\tabcolsep}{2.5pt} 
\resizebox{\linewidth}{!}{ 
\begin{tabular}{l|cc|cc|cc|cc}
\hline
\multirow{2}{*}{Method} & \multicolumn{2}{c|}{CIF10} & \multicolumn{2}{c|}{CIF100} & \multicolumn{2}{c|}{LT-CIF10} & \multicolumn{2}{c}{LT-CIF100} \\ \cline{2-9} 
 & ACC$\uparrow$ & AUC$\uparrow$ & ACC$\uparrow$ & AUC$\uparrow$ & ACC$\uparrow$ & AUC$\uparrow$ & ACC$\uparrow$ & AUC$\uparrow$ \\ \hline

OE (Baseline) & 93.12 $\pm$ 0.05 & 95.41 $\pm$ 0.02 & 73.86 $\pm$ 0.10 & 81.96 $\pm$ 0.12 & 69.73 $\pm$ 0.12 & 85.11 $\pm$ 0.06 & 38.93 $\pm$ 0.07 & 68.61 $\pm$ 0.05 \\ \hline

+ SKD  & 93.39 $\pm$ 0.09 & 96.47 $\pm$ 0.05 & 74.20 $\pm$ 0.08 & 84.56 $\pm$ 0.30 & 74.79 $\pm$ 0.12 & 90.10 $\pm$ 0.04 & 39.59 $\pm$ 0.11 & 72.80 $\pm$ 0.08 \\
+ SOS  & 92.90 $\pm$ 0.04 & 95.64 $\pm$ 0.01 & 73.62 $\pm$ 0.11 & 82.90 $\pm$ 0.11 & 71.35 $\pm$ 0.16 & 86.43 $\pm$ 0.05 & 37.77 $\pm$ 0.10 & 69.70 $\pm$ 0.03 \\
+ OSCL & \textbf{93.70} $\pm$ 0.06 & 95.72 $\pm$ 0.04 & \textbf{74.82} $\pm$ 0.08 & 83.28 $\pm$ 0.16 & 72.47 $\pm$ 0.18 & 86.03 $\pm$ 0.10 & \textbf{40.89} $\pm$ 0.07 & 70.75 $\pm$ 0.10 \\ \hline

\textbf{Ours (Combined)}& 93.38 $\pm$ 0.10 & \textbf{96.70} $\pm$ 0.06 & 73.13 $\pm$ 0.19 & \textbf{85.59} $\pm$ 0.22 & \textbf{77.67} $\pm$ 0.11 & \textbf{90.81} $\pm$ 0.08 & 39.73 $\pm$ 0.09 & \textbf{74.84} $\pm$ 0.15 \\ \hline
\end{tabular}
}
\label{D_main}
\end{table}
\subsection{Combined Approach}
\label{combined}
To leverage the individual strengths of the proposed factors, we simultaneously integrate Semi-hard Outlier Sampling (SOS), Self-Knowledge Distillation (SKD), and Outlier-aware Supervised Contrastive Learning (OSCL) into a unified fine-tuning framework. The final objective is formulated by combining the respective regularization terms into the total loss: 
\begin{equation} 
    \small
    \label{equation:total}
L_{total}=L_{cls}+\lambda_{reg}L_{reg}+\lambda_{KD}L_{KD}+\lambda_{SC}L_{SC}.
\end{equation}

As summarized in our incremental ablation study (see Table~\ref{D_main}), this combined approach yields significant cumulative improvements in both OOD detection and classification accuracy.

\begin{table}[t!]
\centering
\setlength{\tabcolsep}{1pt} 
\scriptsize 
 \centering
\caption{Comparison with other methods on CIFAR using ResNet18. (a) Results on CIFAR10; (b) Results on CIFAR100. Abbreviations: E-OE (EnergyOE \cite{liu2020energy}). "100\%" denotes the use of the full TinyImages-300K dataset (300K images), and "(fine)" indicates our implementation of the originally scratch-based algorithm adapted for the fine-tuning paradigm.}
\label{SCOOD_comparison_balanced}

\begin{subtable}{0.48\linewidth}
\centering
\scriptsize
\caption{}
\label{balanced_cifar10}
\begin{tabular}{c|c|c|c|c}
\hline
Method         &ACC$\uparrow$     &FPR$\downarrow$     &      AUC$\uparrow$   &     PR$\uparrow$         \\ \hline
MSP(ST)         & 93.69          & 31.32          & 89.25         & 86.63          \\
 Energy(ST)      & 93.69          & \textbf{29.07}          & 91.55         & 89.88          \\
 ReAct(ST)~\cite{sun2021react}  & \textbf{93.70} & 29.18 & 91.56 & 89.90  \\
 kNN(ST)~\cite{sun2022out}  & 93.69 & 47.45 & \textbf{91.67} & \textbf{90.78}  \\ \hline
  OE(scratch)       & 90.12          & 20.97          & 95.35         & 95.22          \\ 
   E-OE(scratch) & 90.50          & \textbf{17.91}         & \textbf{95.85}          & \textbf{95.78} \\
      WOOD(100\%) \cite{katz2022training}     & 93.20          &    22.01      &   94.16       &      92.86     \\  
      NTOM(100\%) \cite{chen2021atom}  & \textbf{94.39}           &    22.24      &   92.49       &      90.61     \\  \hline
  VOS \cite{du2022vos}    & \textbf{94.98}           &    42.87     &  90.22    &      90.11    \\  
    NPOS \cite{tao2023non}    & -           &   61.94     &  88.55   &      85.28    \\   
    
    DivOE \cite{zhu2023diversified}  & 93.06     &   18.70     &   95.48   &    95.08     \\
    DivOE(100\%) \cite{zhu2023diversified}  & 93.40        &    17.83  &   95.83    &    95.57    \\

    DivMix\cite{yao2024out}  & 92.95     &    16.15      &   96.17   &   96.08   \\
    DivMix(100\%)\cite{yao2024out}  & 93.89          &    \textbf{13.84}     &   \textbf{96.72}    &   \textbf{96.66}    \\

    DivMix(fine) \cite{yao2024out}  & 93.68          &    18.97     &   95.39     &    94.88   \\

 \hline 

    OE       & 93.12 & 18.82 & 95.41 & 95.00          \\ 
        \textbf{OE+Ours}           & \textbf{93.38} & \textbf{13.81} & \textbf{96.70} & \textbf{96.31}          \\ \hline
        E-OE        & 93.33 & 14.45 & 96.81 & 96.73          \\ 
        \textbf{E-OE+Ours}           & \textbf{93.64} & \textbf{13.15} & \textbf{96.99} & \textbf{96.84}          \\ \hline
        OECC       & 91.81 & 14.28 & 96.40 & 95.44          \\ 
        \textbf{OECC+Ours}           & \textbf{92.18} & \textbf{13.06} & \textbf{96.52} & \textbf{95.66}          \\ \hline
       \end{tabular}
\end{subtable}
\begin{subtable}{0.48\linewidth}
\centering
\scriptsize
\caption{}
\begin{tabular}{c|c|c|c|c}
\hline
Method         &ACC$\uparrow$     &FPR$\downarrow$     &      AUC$\uparrow$   &     PR$\uparrow$         \\ \hline
MSP(ST)         &  \textbf{75.70}          & 62.78          & 76.14         & 71.29          \\
 Energy(ST)      & \textbf{75.70}          & 57.59          & 79.78         & 73.31          \\
 ReAct(ST)~\cite{sun2021react}  & 74.79 & \textbf{55.12} & \textbf{80.90} & 75.35  \\
 kNN(ST)~\cite{sun2022out}  & \textbf{75.70} & 76.65 & 79.13 & \textbf{76.86}  \\ \hline
  OE(scratch)       & 66.47          & 61.50          & 80.71         & 77.44          \\ 
   E-OE(scratch) & 65.92          & 55.67          & 81.54          & 76.82 \\
   WOOD(100\%) \cite{katz2022training}    & \textbf{75.93}           &    \textbf{51.52}      &   \textbf{82.97}       &      \textbf{77.66}     \\   
      NTOM(100\%) \cite{chen2021atom}  & 72.41           &    61.77      &   75.58       &    70.71      \\ \hline
   
   VOS \cite{du2022vos}     & \textbf{77.44}           &     55.90      &    80.41      &       74.02     \\ 
    NPOS \cite{tao2023non}  & -         &   79.69      &  75.06       &   70.19     \\ 
    DivOE \cite{zhu2023diversified}  & 74.05        &    53.75     &  81.87  &    77.13     \\
    DivOE(100\%) \cite{zhu2023diversified}  & 74.97          &    50.90    &   83.26     &    78.47     \\

    DivMix\cite{yao2024out}  & 72.49           &    49.01      &  83.09    &   78.12    \\
    DivMix(100\%)\cite{yao2024out}  & 73.63         &    \textbf{45.45}      &   \textbf{84.71}     &    \textbf{79.87}    \\
    DivMix(fine) \cite{yao2024out}  & 75.79       &   51.84     &  82.49     &   77.39   \\ \hline

    OE        & \textbf{73.86} & 53.38 & 81.96 & 76.95          \\ 
        \textbf{OE+Ours}           & 73.13 & \textbf{43.16} & \textbf{85.59}& \textbf{80.44}          \\ \hline
        E-OE      & 74.50 & 43.59 & 86.07 & 81.34          \\ 
        \textbf{E-OE+Ours}           & \textbf{74.93} & \textbf{42.46} & \textbf{86.38} & \textbf{81.49}         \\ \hline
        OECC       & 69.36 & 45.41 & 84.01 & 77.84          \\ 
        \textbf{OECC+Ours}           & \textbf{72.23} & \textbf{41.78} & \textbf{85.64} & \textbf{80.26}          \\ \hline
       \end{tabular}
\label{balanced_cifar100}
\end{subtable}
\label{SCOOD_comparison_balanced}
\end{table}

\section{Experimental Result}
\label{exp_section}
\subsection{Experiment Settings}
\label{exp_setting}
\noindent\textbf{In-distribution (ID) Dataset.} 
We use CIFAR \cite{krizhevsky2009learning} and long-tailed CIFAR \cite{cao2019learning} as ID datasets. We denote balanced CIFAR datasets as CIF and long-tailed CIFAR as LT-CIF.
We also use LT-ImageNet~\cite{liu2019large} for large-scale long-tailed OOD detection.

\noindent\textbf{Auxiliary Dataset.}
We use TinyImages-300K dataset~\cite{hendrycks2018deep}.
For LT-ImageNet OOD detection, we use ImageNet-Extra~\cite{wang2022partial} and follow PASCL \cite{wang2022partial} settings.
\noindent\textbf{Out-of-distribution (OOD) Dataset.} 
We evaluate the proposed method for various OOD datasets. We mainly use Texture \cite{cimpoi2014describing}, SVHN \cite{netzer2011reading}, TinyImageNet \cite{le2015tiny}, LSUN \cite{yu2015lsun}, Places365 \cite{zhou2017places}, and CIFAR \cite{krizhevsky2009learning}. We denote these OOD datasets as SC-OOD benchmark. Note that if a model is trained on CIFAR10, then OOD CIFAR dataset is CIFAR100, and vice versa. We also evaluate our algorithm to other OOD datasets called MOOD \cite{lin2021mood} benchmark that is composed of 11 datasets. The details of OOD benchmarks are described in the appendix.

\noindent\textbf{Hyperparameter Settings.}
 We use ResNet-18~\cite{he2016deep} (Generalization to WideResNet~\cite{zagoruyko2016wide} in appendix). Base scratch models are trained for 200 epochs using Adam with $LR=0.001$ and weight decay $0.0005$, following PASCL~\cite{wang2022partial} parameters and OE~\cite{hendrycks2018deep} augmentation. For fine-tuning the pre-trained model, we set a learning rate of 0.001, weight decay of 0.0005, in-distribution batch size of 128, auxiliary batch size of 256, cosine annealing \cite{loshchilov2016sgdr} scheduler, and SGD optimizer. When ID dataset is CIFAR \cite{krizhevsky2009learning} and LT-CIFAR \cite{cao2019learning}, we fine-tune the model for 10 epochs and 20 epochs, respectively. We set $\lambda_{reg}=5, \lambda_{KD}=1, \lambda_{SC}=1$ for all experiments. Dataset-specific hyperparameters $\{T_{KD}, q, n\}$ are set to $\{4, 0.75, 8\}$ for CIFAR10, $\{4, 0.85, 8\}$ for CIFAR100, $\{4, 0.8, 4\}$ for LT-CIFAR10, and $\{4, 0.9, 8\}$ for LT-CIFAR100. All results are averaged over 8 random seeds. The training details is presented in the appendix.
\subsection{Comparison with other methods}

\begin{table}[t!]

\caption{ Comparison with other methods on LT-CIFAR using ResNet18. (a): Results on LT-CIFAR10; (b): Results on LT-CIFAR100. Abbreviations: E-OE (EnergyOE \cite{liu2020energy}), BE-OE (Balanced EnergyOE \cite{choi2023}), and (Po) denotes the application of Post-processing as used in COCL \cite{miao2024out}}.

\begin{subtable}{0.48\linewidth}
\centering
\scriptsize

\caption{}
\label{imbalanced_cifar10}
\begin{tabular}{c|c|c|c|c}
\hline
Method         &ACC$\uparrow$     &FPR$\downarrow$     &      AUC$\uparrow$   &     PR$\uparrow$         \\ \hline
MSP(ST)         & \textbf{73.28}          & 61.30          & 74.55     & 72.26          \\
 Energy(ST)      & \textbf{73.28}         & \textbf{53.82}   & \textbf{80.33}    & 77.02          \\
 ReAct(ST)~\cite{sun2021react}  & \textbf{73.28} & 53.83 & \textbf{80.33} & 77.02  \\
 kNN(ST)~\cite{sun2022out}  & \textbf{73.28} & 74.98 & 80.11 & \textbf{77.48}  \\ \hline
  OE(scratch)       & 72.85          & 34.79          & 89.40         & 85.79          \\ 
   E-OE(scratch) & 73.40          & 34.36        & 86.52          & 81.56 \\
   PASCL~\cite{wang2022partial}     &  77.08           &    33.60     &   90.72     &    \textbf{88.89}     \\    
   OS~\cite{wei2022open}   & 77.04           &   \textbf{30.74}        &  90.33        & 85.61       \\ 
   COCL~\cite{miao2024out}     &  73.38           &   36.75    &   90.56    &    88.85     \\     
   COCL(Po)~\cite{miao2024out}     &  \textbf{78.67}        &   32.18     &  \textbf{91.15}      &    88.88     \\     \hline   
   VOS~\cite{du2022vos}     &  68.76        &    71.14  &  70.72    &    69.43    \\    
      NPOS~\cite{tao2023non}     &  -          &    98.15     &   44.32     &    43.2     \\    
    DivOE \cite{zhu2023diversified}  & 69.10           &    51.27     &   85.25     &    84.69     \\
    DivOE(100\%) \cite{zhu2023diversified}  & 69.10           &    51.28      &   85.25     &    84.69     \\
    DivMix\cite{yao2024out}  & 59.34          &    51.87     &  85.09   &   84.81    \\
    DivMix(100\%)\cite{yao2024out}  & 58.64          &    53.84  &   84.34   &   84.05    \\
   DivMix(fine)~\cite{yao2024out}     &  \textbf{73.66}        &    \textbf{46.77}    & \textbf{86.55}     &    \textbf{86.34}    \\     \hline  

    OE        & 69.73 & 51.51 & 85.11 & 84.59         \\ 
        \textbf{OE+Ours}           &  77.67 &  37.19 &  \textbf{90.81} &  \textbf{90.50}      \\ 
        \textbf{OE+Ours(Po)}           &  \textbf{79.08} &  \textbf{35.47} &  90.75 &  88.92      \\

        \hline
        E-OE       & 74.77 & 33.74 & 91.86 & 91.91          \\ 
        \textbf{E-OE+Ours}           & 78.18 &  32.15 & 92.10 &  92.01          \\
        \textbf{E-OE+Ours(Po)}           & \textbf{80.24} & \textbf{29.55} & \textbf{92.58} &  \textbf{92.24}          \\
        
        \hline
        OECC        & 62.01 & 44.52 & 87.86 & 87.56          \\ 
        \textbf{OECC+Ours}           & \textbf{ 75.23}&  \textbf{32.86} &  \textbf{91.53} &  \textbf{90.49}          \\\hline 
        BE-OE       & 76.30 & 30.94 & 92.52 &  \textbf{91.86}          \\ 
        \textbf{BE-OE   +Ours}           &  \textbf{78.13} & \textbf{29.00} & \textbf{92.57} & 91.50          \\         
        \hline
       \end{tabular}
\end{subtable}
\begin{subtable}{0.48\linewidth}
\centering
\scriptsize

\caption{}
\begin{tabular}{c|c|c|c|c}
\hline
Method         &ACC$\uparrow$     &FPR$\downarrow$     &      AUC$\uparrow$   &     PR$\uparrow$         \\ \hline
MSP(ST)         & \textbf{40.22}         & 83.30         & 61.17     & 58.10          \\
 Energy(ST)      & \textbf{40.22}          & 80.59      & 64.08        & 59.86          \\
 ReAct(ST)~\cite{sun2021react}  & 39.02 & \textbf{79.72} & \textbf{64.22} & 59.90  \\
 kNN(ST)~\cite{sun2022out}  & \textbf{40.22} & 87.97 & 62.91 & \textbf{60.41}  \\ \hline
  OE(scratch)       & 41.31          & 69.93          & 73.32         & 67.92          \\ 
   E-OE(scratch) & 40.78          & 67.46        & 74.30          &  \textbf{70.09}\\
   PASCL~\cite{wang2022partial}    &  43.10          &    67.51    &  73.40     & 67.02       \\   
      OS~\cite{wei2022open}  & 39.96           &     \textbf{66.84}      &  \textbf{74.39}       &  69.37     \\   
   COCL~\cite{miao2024out}     &  41.5           &   70.04    &   73.69      &    69.35    \\     
   COCL(Po)~\cite{miao2024out}     &  \textbf{44.10}          &  67.54  &  74.23     &    69.19     \\   \hline   
   VOS~\cite{du2022vos}     &  37.71         &   81.39     &  63.01    &   59.05   \\ 
     NPOS~\cite{tao2023non}     &  -          &   96.20     &   56.22      &    53.00     \\     
    DivOE \cite{zhu2023diversified}  & 38.92          &    76.59   &   68.65   &    64.76    \\
    DivOE(100\%) \cite{zhu2023diversified}  & 38.92     &    76.59    &  68.65    &    64.76     \\
    DivMix\cite{yao2024out}  & 36.35          &   74.64     &  69.98   &    66.77    \\
    DivMix(100\%)\cite{yao2024out}  & 36.35         &    74.64    &   69.98   &    66.77   \\ 
   DivMix(fine)~\cite{yao2024out}     &  \textbf{40.43}           &   \textbf{71.15}     &   \textbf{73.37}     &    \textbf{69.13}     \\  \hline      

    OE       & 38.93 & 76.87 & 68.61 & 64.78          \\ 
        \textbf{OE+Ours}           &  39.73 &  67.95 &  74.84 &  70.81          \\
         \textbf{OE+Ours(Po)}           &  \textbf{41.91}& \textbf{64.88} & \textbf{76.42} &  \textbf{72.28}     \\ 

        \hline
        E-OE     & 40.54 & 64.54 & 76.33 & 72.13          \\ 
        \textbf{E-OE+Ours}           &  40.76 & 60.34 & 77.50 &  73.17          \\ 
            \textbf{E-OE+Ours(Po)}           &  \textbf{43.66} & \textbf{59.27} & \textbf{77.95} &  \textbf{73.49}       \\ 

    \hline
        OECC        & 31.08 & 73.96 & 71.70 & 68.42          \\ 
        \textbf{OECC+Ours}           &  \textbf{39.77} &  \textbf{64.04} &  \textbf{75.91} &  \textbf{71.23}          \\\hline 
        BE-OE     & 40.78 & 61.30 & 77.64 & 73.00          \\ 
        \textbf{BE-OE   +Ours}           & \textbf{43.45} &  \textbf{60.90} & \textbf{78.01} & \textbf{73.50}          \\         
        \hline
       \end{tabular}
\label{imbalanced_cifar100}
\end{subtable}
\label{SCOOD_comparison_imbalanced}
\end{table}

\subsubsection{Balanced dataset}
As summarized in Table~\ref{SCOOD_comparison_balanced}, we compare our approach with various state-of-the-art methods on balanced CIFAR datasets. For a comprehensive analysis, we report results using both the full TinyImages-300K dataset (100\%) and a 30K subset (10\%), the latter of which follows the setting of our proposed method. Since DivMix~\cite{yao2024out} was originally designed for training from scratch, we also implemented and evaluated a fine-tuned version, DivMix(fine), to ensure a fair comparison within the fine-tuning paradigm.

We observe two key insights from the empirical results. First, synthetic-outlier-based methods such as VOS~\cite{du2022vos} and NPOS~\cite{tao2023non} generally underperform compared to real-outlier-based approaches like DivOE~\cite{zhu2023diversified}, DivMix~\cite{yao2024out}, and our proposed framework. For instance, on CIFAR10 (Table~\ref{balanced_cifar10}), VOS and NPOS achieve AUROCs of 90.22\% and 88.55\%, respectively, which are significantly lower than the 96.70\% achieved by OE+Ours. This highlights that leveraging actual outlier distributions provides a more robust foundation for OOD detection than relying on synthetic generation.

Second, our method achieves superior or comparable performance to specialized real-outlier enhancement methods (DivOE, DivMix) without requiring any additional outlier data augmentation. Notably, on CIFAR100 (Table~\ref{balanced_cifar100}), OE+Ours reaches an AUROC of 85.59\% using only 10\% of the outlier data, outperforming DivMix(100\%) which utilizes the full 300K dataset (84.71\%). These findings demonstrate that our approach effectively improves OOD detection and classification accuracy through simple yet orthogonal algorithmic factors, underscoring its efficiency and practicality in real-world scenarios.

\subsubsection{Imbalanced dataset}
The empirical results on imbalanced datasets summarized in Table~\ref{SCOOD_comparison_imbalanced}, provide several critical insights into the robustness of OOD detection under label distribution shifts.
First, we observe a significant performance collapse in synthetic-outlier-based methods (VOS, NPOS) and scratch-training methods (DivMix) when applied to imbalanced scenarios. While these methods show promise in balanced settings, they fail to maintain OOD discriminative power in long-tailed distributions; for instance, NPOS drops to an AUROC of 44.32\% on LT-CIFAR10. In contrast, fine-tuning-based approaches demonstrate remarkable robustness, successfully extending their OOD detection capabilities to imbalanced scenarios without the catastrophic accuracy loss seen in scratch-trained models like DivMix, which drops to 58.64\% accuracy.

Second, methods relying on outlier data augmentation (DivOE, DivMix) struggle to provide robust improvements in imbalanced settings. DivOE shows only marginal gains over the standard OE baseline, while DivMix actually underperforms its base EnergyOE model. This suggests that simply augmenting outliers is insufficient to overcome the challenges posed by ID class imbalance.

Third, our approach remains highly competitive and powerful in these challenging scenarios. To ensure a fair comparison with recent imbalanced-specific methods like COCL~\cite{miao2024out}, we integrated the post-processing (Po) technique—originally proposed for COCL—into our framework. As shown in Table~\ref{imbalanced_cifar10} and \ref{imbalanced_cifar100}, our method with post-processing (\textbf{OE+Ours(Po)}) significantly outperforms COCL(Po) in both accuracy and OOD detection metrics. While methods such as PASCL, OS, and COCL require specialized training from scratch for imbalanced data, our results reaffirm that a well-regularized fine-tuning approach is not only simpler but also a superior paradigm for practical OOD detection in long-tailed environments.

\subsubsection{Large-Scale Imbalanced dataset}


\begin{table}[t!]

\caption{Comparison result on LT-ImageNet using ResNet50. (a): Result on each LT-ImageNet benchmark dataset (b): Comparison result on LT-ImageNet benchmark (over 5 datasets).}
\label{imagenet_res}
\begin{subtable}{0.48\linewidth}
\centering
\scriptsize

\caption{}
\label{imagenet_res_a}
\begin{tabular}{c|c|c|c|c}
\hline
Dataset  & Method & FPR$\downarrow$ & AUC$\uparrow$ & PR$\uparrow$ \\ \hline
\multirow{2}{*}{ImageNet OOD}  & OE  &  86.25   &  70.39 & 72.90   \\
& \textbf{OE+Ours}    & \textbf{85.55} & \textbf{72.65} & \textbf{75.84}   \\ \hline
\multirow{2}{*}{iNaturalist}  & OE    &  48.58   & 89.01    & 70.18   \\
& \textbf{OE+Ours}    & \textbf{41.66}  & \textbf{91.50}    & \textbf{74.19}  \\ \hline
\multirow{2}{*}{SUN}  & OE    & 76.69    &  74.02   &   40.37 \\
& \textbf{OE+Ours}    & \textbf{72.68}    & \textbf{78.34}    & \textbf{46.02}   \\ \hline
\multirow{2}{*}{Places}  & OE    &  77.43   &  71.82   & 34.51   \\
& \textbf{OE+Ours}    & \textbf{74.14}    & \textbf{75.27}    &  \textbf{38.11}  \\ \hline
\multirow{2}{*}{DTD}  & OE    &  78.30   & 70.10    & 20.63   \\
& \textbf{OE+Ours}    &  \textbf{73.89}   & \textbf{79.81}    & \textbf{40.64}   \\ \hline
\multirow{2}{*}{Average}  & OE    &  73.45   & 75.07    & 47.72   \\
& \textbf{OE+Ours}    & \textbf{69.58}    & \textbf{79.51}    & \textbf{54.96}   \\ \hline
\end{tabular}
\end{subtable}
\begin{subtable}{0.48\linewidth}
\centering
\scriptsize

\caption{}
\label{imagenet_res_b}
\begin{tabular}{c|c|c|c|c}
\hline

Method  &ACC$\uparrow$   &FPR$\downarrow$     &      AUC$\uparrow$   &     PR$\uparrow$         \\ \hline

    MSP(ST)     & \textbf{40.80} & 80.95 & 66.78 & 34.32  \\

    Energy(ST)     & \textbf{40.80} & \textbf{70.34} & 75.03 & 45.76  \\

    OE(scratch)     & 40.17 & 73.42 & 76.38 & 49.53  \\

    PASCL~\cite{wang2022partial}     & 40.25 & 72.75 & \textbf{76.48} &  \textbf{50.02} \\ \hline
    
    OE     & 38.46 & 73.45 & 75.07 & 47.72  \\
    
    \textbf{OE+Ours}     & \textbf{39.20} & \textbf{69.58} & \textbf{79.51} & \textbf{54.96}  \\ \hline
    
\end{tabular}
\end{subtable}
\label{MOOD_table_all}
\end{table}

We validate our method on Large-Scale dataset ImageNet-LT using ResNet50.
We follow the protocol of PASCL~\cite{wang2022partial}.
In this setting, we utilize full data of training outlier same as PASCL.
Thus, we only adopt SKD and OSCL excluding SOS.
As in Table~\ref{imagenet_res_a}, Our method improves the OOD detection performance of OE by a large margin. 
As summarized in Table~\ref{imagenet_res_b},
our method surpasses recent approaches such as PASCL in terms of OOD detection AUROC.

\subsection{Comparison result on other OOD benchmark}


\begin{table}[t!]

\caption{Comparison result on LT-CIFAR10 using ResNet18; Mean over 8 random runs are reported; OOD detection average performance and accuracy (a): Result on MOOD benchmark (over 11 datasets) (b): Result on SYN benchmark (over 3 datasets).}
\begin{subtable}{0.48\linewidth}
\centering
\scriptsize
\caption{}
\label{MOOD_lt_cifar10}
\begin{tabular}{c|c|c|c|c}
\hline
Method         & ACC$\uparrow$ & FPR$\downarrow$ & AUC$\uparrow$ & PR$\uparrow$ \\ \hline
OE         & 69.73    &  40.71   &  87.39   &  85.48  \\ 
\textbf{OE+Ours}        &  \textbf{77.66}   &  \textbf{30.46}   & \textbf{91.14}    &  \textbf{89.88}  \\ \hline
E-OE     &  74.77   & 31.22    &  90.99   & \textbf{89.87}   \\ 
\textbf{E-OE+Ours}  &  \textbf{78.18}   & \textbf{28.67}    & \textbf{91.29}    & 89.73   \\ \hline
OECC          & 62.01    &  34.78   &  89.48   & \textbf{87.94}   \\ 
\textbf{OECC+Ours}      &  \textbf{75.23}   &  \textbf{30.06}   &  \textbf{90.53}   & 87.73   \\ \hline
BE-OE     &  76.30   & 27.59    &  91.91   & \textbf{90.46}   \\ 
\textbf{BE-OE+Ours} &  \textbf{78.13}   &  \textbf{25.99}   &  \textbf{92.09}   &  \textbf{90.46}  \\ \hline
\end{tabular}
\end{subtable}
\begin{subtable}{0.48\linewidth}
\centering
\scriptsize
\caption{}
\label{syn_imbalanced_cifar10}
\begin{tabular}{c|c|c|c|c}
\hline

Method         &ACC$\uparrow$     &FPR$\downarrow$     &      AUC$\uparrow$   &     PR$\uparrow$         \\ \hline

    OE     & 69.73 & 30.27 & 80.03 & 77.69    \\
    
        \textbf{OE+Ours}           & \textbf{77.66} & \textbf{20.90} & \textbf{92.32} & \textbf{87.79}     \\ \hline
        
        E-OE    & 74.77 & 9.67 & 96.60 & 94.26 \\ 
        
        \textbf{E-OE+Ours}    & \textbf{78.18} &\textbf{7.79}& \textbf{97.35} &\textbf{95.01}     \\ \hline
        
        OECC        & 62.01 & 21.69 & 90.80 & 87.84   \\ 
        
        \textbf{OECC+Ours}     & \textbf{75.23} & \textbf{15.49} & \textbf{94.96} & \textbf{91.70}     \\\hline 
        
        BE-OE & 76.30 & 7.52 & 97.51 & 95.23  \\ 
        
        \textbf{BE-OE+Ours}   & \textbf{78.13} & \textbf{6.85} & \textbf{97.81} & \textbf{95.58}          \\         
        \hline
       \end{tabular}
\end{subtable}
\label{MOOD_table_all}
\vspace{-0.3cm}
\end{table}

To validate the generality of our approach across various OOD test distributions, we evaluate our result on other OOD test sets which are MOOD and SYN benchmarks.
We summarize our comparison result on LT-CIFAR10 in Table~\ref{MOOD_table_all}.
Similarly, as SC-OOD benchmark, Our approach outperformed fine-tuning-based algorithms, in terms of accuracy and OOD detection. We observe similar trends on other in-distribution datasets (i.e. CIFAR100). Comprehensive comparisons, along with results for different in-distribution datasets on both the MOOD benchmark and SYN benchmark, are provided in the supplemental material.
\subsection{Comparison with other contrastive loss}

\begin{table}[h!]
\caption{Result of applying Outlier-aware Supervised Contrastive Learning to existing methods; $n$ is set to 1 for existing methods.}
\centering
\scriptsize
\label{C_comparison_result}
\begin{tabular}{c|cccc}
\cline{2-5}
     & \multicolumn{4}{c}{AUC $\uparrow$}                     \\ \cline{1-5} 

Method     & CIFAR10 & CIFAR100 & LT-CIFAR10 & LT-CIFAR100  \\ \hline
base OE           &  95.41       &    81.96      & 85.11  &   68.61          \\ \hline

SCL~\cite{khosla2020supervised}           &  95.44       &    82.44      & 85.04  &   69.02          \\
\textbf{SCL+OSCL}           &   \textbf{95.72}     &    \textbf{83.28} &   \textbf{86.03}    &   \textbf{70.75}           \\ \hline

CIDER~\cite{ming2022cider}           & 95.44   &    82.43  &  84.98  &  69.03       \\
\textbf{CIDER+OSCL}         &  \textbf{95.72}      &   \textbf{83.28}   &  \textbf{85.99}     &  \textbf{70.75}           \\ \hline

BCL~\cite{zhu2022balanced}      &  95.44    &  82.44  & 84.93     &   69.01         \\
\textbf{BCL+OSCL}           & \textbf{95.72}      &   \textbf{83.30}      &  \textbf{85.96}       &   \textbf{70.77}         \\ \hline
\end{tabular}

\end{table}
\label{C_experiment}
We enhance the traditional supervised contrastive loss by incorporating outlier data as negative samples and utilizing multi-batch transform. The setting of $n$ is described in Section~\ref{exp_setting}. Our OSCL approach outperforms existing methods, both in balanced datasets like CIFAR and imbalanced datasets like LT-CIFAR, demonstrating its effectiveness across different dataset distributions. The results in Table~\ref{C_comparison_result} highlight the impact of our approach during the fine-tuning process.

\section{Analysis}
\subsection{How OE effects on ID accuracy}
\label{analysis_accuracy}
\begin{table}[h!]
\vspace{-0.3cm}

\caption{Analysis of ID classification accuracy on LT-CIFAR10 as ID data   (Energy value or cardinality $\times 10^3$; scaling). Result on a test set of CIFAR10 with 10000 samples.
$E_{i}$ is the sum of energy for entities predicted as class $i$.
$E_{i,j}$ is sum of $E_{i}$ and $E_{j}$. Class 0 to 4 belongs to the Head class and class 5 to 9 belongs to the Tail class. Cardinality represents the number of prediction labels.}
\centering
\scriptsize
\label{ana_accuracy}
\vspace{-0.3cm}

\begin{subtable}{1.0\linewidth}
\centering
\scriptsize
\caption{}


\begin{tabular}{c|cccccc}
\hline
 & $-E_{0, 1}$ & $-E_{2, 3}$ & $-E_{4,5}$ & $-E_{6,7}$ & $-E_{8,9}$  & Ratio \\ \hline
Base & 29.42 & 20.17 & 14.76 & 12.49 & 10.28 & 1.86  \\ \hline
\multirow{2}{*}{OE} & 14.79 & 8.02 & 5.93 & 3.50 & 1.35 & \multirow{2}{*}{9.92}  \\ 
& (-14.64) & (-12.14) & (-8.84) & (-8.99) & (-8.92) &   \\ \hline
\multirow{2}{*}{\textbf{OE+SKD}} & 24.54 & 14.02 & 10.87 & 8.30 & 4.54 & \multirow{2}{*}{4.41}  \\ 
& (-4.89) & (-6.15) & (-3.89) & (-4.19) & (-5.74) &   \\ \hline
\end{tabular}

\end{subtable}

\begin{subtable}{1.0\linewidth}
\centering
\scriptsize
\caption{}

\begin{tabular}{c|cc|cc|ccc}
\hline
\multirow{2}{*}{} & \multicolumn{2}{c|} {Neg Energy}  & \multicolumn{2}{c|} {Cardinality} & \multicolumn{3}{c} {ACC$\uparrow$}  \\ \cline{2-8}
& Head & Tail & Head & Tail & Total & Head & Tail \\ \hline

Base & 57.80 & 29.33 & 6.71 & 3.29 & 73.28 & 86.14 & 60.42   \\ \hline
OE only & 26.72 & 6.88 & 7.36 & 2.64 & 65.90 & 85.46 & 46.34   \\ \hline
SKD only & 57.81 & 30.58 & 6.66 & 3.34 & 73.90 & \textbf{86.40} & 61.40   \\ \hline
\textbf{OE+SKD} & 45.14 & 17.13 & 6.50 & 3.50 & \textbf{74.74} & 86.36 & \textbf{63.12}  \\ \hline

\end{tabular}
\end{subtable}

\end{table}


Energy in EnergyOE~\cite{liu2020energy} is the negative value of logsumexp over logits, making negative energy the logsumexp over logits. We analyze prediction tendencies through the sum of predicted negative energy for each class.
Compared to the base model, the OE model similarly suppresses negative energy for each class, but it worsens the relative prediction imbalance between head and tail classes. As seen in Table~\ref{ana_accuracy}, the relative ratio between $E_{0,1}$ and $E_{8,9}$ increases from 1.86 to 9.92. Our SKD method alleviates this by learning the original distribution, resulting in a relative ratio of 4.41.
In summary, OE suppresses negative energy for in-distribution (ID) data, intensifying the imbalance between head and tail classes and significantly reducing accuracy for tail classes. Our SKD approach mitigates this, maintaining good accuracy for tail classes.


\subsection{How SKD improves OOD detection}
\begin{table}[h!]

\caption{Analysis OOD detection performance. LT-CIFAR10 as ID data and Texture as OOD test data.}
\centering
\scriptsize

\begin{tabular}{c|ccc}
\hline

 & ID neg. energy (avg) & OOD neg. energy (avg) & AUC$\uparrow$  \\ \hline
Base & 8.71  & 6.05 (-2.66) & 65.74   \\ \hline
OE only & 3.36 &  2.34 (-1.02) & 73.44    \\ \hline
SKD only & 8.84 & 6.12 (-2.72) & 65.97     \\ \hline
\textbf{OE+SKD} & 6.23 & 2.45 (-3.78) & \textbf{79.40}    \\ \hline

\end{tabular}

\end{table}
\label{analysis_OOD}
OE with SKD not only boosts accuracy but also outperforms the previous OE approach in OOD detection. The gap in negative energy between ID and OOD illustrates this effect. While OE alone enhances OOD performance by reducing OOD energy, it simultaneously lowers the energy for actual ID, resulting in a negligible difference in average negative energy between ID and OOD. In contrast, OE+SKD maintains the original negative energy for ID, accentuating the difference between ID and OOD and improving OOD detection.

\subsection{OSCL improves ID-OOD seperability}
\label{analysis_OSCL}
\begin{table}[h!]
\centering
\setlength{\tabcolsep}{2pt} 
\scriptsize

\caption{Analysis of OSCL. L(C) means LSUN(crop). SCL is from \cite{khosla2020supervised}.}
\label{ana_scl}

\vspace{-0.2cm}

\begin{subtable}{0.38\linewidth}
\centering
\caption{}
\label{ana_scl_a}
\begin{tabular}{l|cc}
\hline
Method & Disp.$\uparrow$ & Comp.$\downarrow$ \\ \hline
OE only & 73.65 & 17.80 \\
OE+SCL & 75.35 & 16.53 \\
\textbf{OE+OSCL} & \textbf{77.36} & \textbf{16.38} \\ \hline
\end{tabular}
\end{subtable}
\hfill 
\begin{subtable}{0.60\linewidth}
\centering
\caption{}
\label{ana_scl_b}
\begin{tabular}{l|ccccc|c}
\hline
\multirow{2}{*}{Method} & \multicolumn{5}{c|}{Sep.$\uparrow$ ($^\circ$)} & \\ \cline{2-7}
 & DTD & SVHN & C100 & L(C) & iSUN & Avg \\ \hline
OE only & 30.02 & 32.44 & 19.11 & 35.01 & 29.68 & 29.25 \\
OE+SCL & 29.99 & 32.33 & 19.76 & 35.56 & 30.65 & 29.66 \\
\textbf{OE+OSCL} & \textbf{35.68} & \textbf{36.92} & \textbf{21.26} & \textbf{42.11} & \textbf{35.50} & \textbf{34.17} \\ \hline
\end{tabular}
\end{subtable}

\end{table}
Our OSCL stands out by using outlier data as negative samples, pushing the contrasting embeddings farther from the ID samples. CIDER~\cite{ming2022cider} evaluates normalized embedding features for metrics like ID-OOD Separability, Dispersion, and Compactness. As seen in Table~\ref{ana_scl}, OSCL excels in promoting robust ID-OOD separation compared to traditional SCL. Additionally, our method exhibits good Compactness and Dispersion in feature embedding. Consequently, OSCL showcases enhanced OOD detection performance over previous SCL.
\section{Ablation Study}

\subsection{Self-Knowledge Distillation effect depending on regularization coefficient}
\label{A_discussion}

\begin{table}[h!]

\caption{Effect of using Self-Knowledge Distillation depending on regularization coefficient $\lambda_{reg}$.}
\centering
\scriptsize
\label{A_reg_coef_ablation}
\begin{tabular}{cc|cc|cc}
\cline{3-6}
\multicolumn{2}{c|}{\multirow{2}{*}{}} &
  \multicolumn{2}{c|}{ACC $\uparrow$} &
  \multicolumn{2}{c}{AUC $\uparrow$} \\ \cline{1-6} 

\multicolumn{1}{c|}{Method} &
  $\lambda_{reg}$  &
  CIFAR10 &
  CIFAR100 &
  CIFAR10 &
  CIFAR100 \\ \hline
  
\multicolumn{1}{c|}{MSP(ST)} & 0.0  & 93.69 & 75.69 & 91.55 & 79.78 \\ \hline
\multicolumn{1}{c|}{\multirow{2}{*}{\begin{tabular}[c]{@{}c@{}}OE\\ (without  SKD)\end{tabular}}} &
  0.5 & \textbf{93.12} 
   & \textbf{73.86} 
   & 95.41
   & 81.96 
   \\
\multicolumn{1}{c|}{}    & \textbf{5.0}  & 92.49  & 72.20  & \textbf{95.81}  & \textbf{82.88}  \\ \hline

\multicolumn{1}{c|}{\multirow{6}{*}{\begin{tabular}[c]{@{}c@{}}OE\\ (wth SKD)\end{tabular}}} &
  0.5 & \textbf{93.73} 
   & \textbf{75.55} 
   & 94.28
   & 80.16 
   \\
   
\multicolumn{1}{c|}{}    & 1.0  & 93.71  & 75.43  & 95.47  & 81.99  \\

\multicolumn{1}{c|}{}    & 3.0  & 93.54  & 74.81   & 96.37 & 84.15  \\

\multicolumn{1}{c|}{}    & \textbf{5.0}  & 93.39  & 74.20  & \textbf{96.47}  & 84.56  \\

\multicolumn{1}{c|}{}    & 7.0  & 93.20  & 73.64   & 96.45  & \textbf{84.64}  \\

\multicolumn{1}{c|}{}    & 10.0 & 92.90 & 73.10 & 96.34 & 84.58  \\ \hline
\end{tabular}

\end{table}

In Table~\ref{A_reg_coef_ablation}, we extensively analyze the impact of Self-Knowledge Distillation. Using OE as the baseline with an initial $\lambda_{reg}$ of 0.5, increasing $\lambda_{reg}$ introduces a trade-off between enhanced out-of-distribution (OOD) performance and reduced accuracy.
We set $\lambda_{KD}=1.0$ and $T_{KD}=4.0$ for self-knowledge distillation loss. As we vary $\lambda_{reg}$ from 0.5 to 10.0, interestingly, setting $\lambda_{reg}$ to 5.0 yields comparable accuracy while significantly improving OOD performance.
Moreover, under the same conditions with $\lambda_{reg}$ set to 5.0, incorporating our self-knowledge distillation consistently demonstrates superior accuracy and OOD detection performance.

\subsection{Supervised Contrastive Learning effect with using n-batch transform}
\label{C_discussion}

\begin{table}[h!]

\caption{Effect of Supervised Contrastive Learning with using $n$-batch transform.}
\centering
\scriptsize
\label{C_ablation_batch}
\begin{tabular}{cc|cc|cc}
\cline{3-6}
\multicolumn{2}{c|}{\multirow{2}{*}{}} &
  \multicolumn{2}{c|}{ACC $\uparrow$} &
  \multicolumn{2}{c}{AUC $\uparrow$} \\ \cline{1-6} 

\multicolumn{1}{c|}{Method} &
  $n$  &
  CIFAR10 &
  CIFAR100 &
  CIFAR10 &
  CIFAR100 \\ \hline
\multicolumn{1}{c|}{MSP(ST)} & 1  & 93.69 & \textbf{75.69}  & 91.55 & 79.78 \\ \hline

\multicolumn{1}{c|}{OE} & 1  & 93.12  & 73.86 & 95.41 & 81.96  \\ \hline

\multicolumn{1}{c|}{\multirow{5}{*}{\begin{tabular}[c]{@{}c@{}}OE \\ +base SCL\end{tabular}}} & 1
   & 93.55 
   & {75.00} 
   & 95.44 
   & 82.44 
   \\
\multicolumn{1}{c|}{}    & 2  & 93.60  & 74.59 & 95.53 & 82.79  \\
\multicolumn{1}{c|}{}    & 4  & 93.64 & 74.64 & 95.59 & 83.08 \\
\multicolumn{1}{c|}{}    & 6  & {93.66} & 74.73 & {95.60} & 83.20  \\
\multicolumn{1}{c|}{}    & 8  & {93.66}  & 74.79  & {95.60} &  {83.26}  \\
\hline

\multicolumn{1}{c|}{\textbf{OE+OSCL} } & 8  & \textbf{93.70}  & 74.82  & \textbf{95.72}  & \textbf{83.28}   \\ \hline
\end{tabular}

\end{table}


Table~\ref{C_ablation_batch} provides the effects of incorporating supervised contrastive learning into the OE framework, as well as the impact of our proposed $n$-batch transform and outlier-aware supervised contrastive loss.
Firstly, when adding the base supervised contrastive loss to the OE framework, we observe simultaneous improvements in both accuracy and OOD detection performance. 
Next, we observe a stepwise improvement in OOD detection performance by gradually increasing the value of $n$ in the $n$-batch transform technique from 1 to 8.
Moreover, by combining the $n$-batch transform technique with our loss, we achieve further improvement compared to the base supervised contrastive loss.

\section{Conclusion}

To address the trade-off between classification accuracy and OOD detection performance in fine-tuning with auxiliary outlier data, we enhance three key elements. By jointly improving these elements, we achieve outstanding accuracy and OOD detection performance compared to four existing fine-tuning-based algorithms.
\section{Acknowledgement}

This work was supported by KAIA/MOLIT [No.2610000618, Development of autonomous driving-based urban environment management service technology].
\bibliographystyle{splncs04}
\bibliography{TIP}
\clearpage
\onecolumn
\section*{\centering
{ \LARGE Supplementary Material}}

\setcounter{equation}{11}
\setcounter{table}{12}
\setcounter{figure}{3}






Our supplementary material covers the following topics:

\noindent \textbf{Societal Impact}: In-depth exploration of our work's broader impacts and limitations, addressing its societal impact.

\noindent \textbf {Datasets}: Detailed explanation of our OOD test benchmark, describing its composition and characteristics.

\noindent \textbf {Implementation Details}: Specifics of our implementation process, including detailed training, hyperparameter tuning, and implementation details for our method and others.

\noindent \textbf {Visualization}: To supplement what couldn't be covered in the main text due to space limitations, we address visualized results depicting sample hardness, and their associated interpretations, and engage in an analysis of the phenomenon posed by extremely hard samples.

\noindent \textbf {Detailed Experimental Results}: 
To supplement the content briefly presented due to space limitations in the main text, we present full results for each OOD test dataset and provide full benchmark comparison tables. These tables encompass results and standard deviations for all in-distribution datasets.

\noindent \textbf {Network Analysis}: 
To supplement the content briefly presented due to space limitations in the main text, we demonstrate a network analysis of our method. We verify our method in WideResNet instead of ResNet18. We provide a full comparison table for the SC-OOD benchmark results conducted using the WideResNet network. This table also includes standard deviations.

\noindent \textbf{ We offer our source code on Github: \\ https://github.com/hyunjunchhoi/Three-factors.}

\section*{\centering
{ \large Societal Impact}}

\subsection{Broader Impacts}
In practical applications, Out-of-Distribution (OOD) detection can be utilized in automated surveillance systems. Furthermore, OOD detection is currently being employed to address anomaly detection in the video beyond just images. Anomaly detection in video sequences mainly aims to obtain positive effects by alarming situations that threaten people's safety. However, the advancement of anomaly detection in the video raises concerns about potential infringement on human privacy and personal freedoms. Therefore, it is important to regulate the usage of these methods to ensure they are employed only in situations where they protect human safety, rather than being used for surveillance or encroaching upon individual privacy.
\subsection{Limitations}
Our method relies on the assumption that there is a sufficient amount of auxiliary data available as outlier data. However, applying our method becomes challenging in environments where obtaining such auxiliary data is extremely difficult or nearly impossible. Moreover, since auxiliary data encompasses a vast range of possibilities from the open world, it is difficult to determine which data truly aligns with the ideal outlier data. Therefore, obtaining clear answers regarding these challenges, such as identifying the ideal outlier data, will be left for future work.
\section*{ \large Datasets}
\subsection{OOD test benchmark}
We evaluate the proposed method for the following three OOD detection benchmarks: SC-OOD~\cite{yang2021semantically}, MOOD~\cite{lin2021mood}, and synthetic OOD (SYN)~\cite{hendrycks2018deep}.

\noindent \textbf {SC-OOD}: 
SC-OOD consists of 6 publicly available datasets: Texture, SVHN, CIFAR, TinyImageNet, LSUN, and Places365. The distinctive aspect is that, among these datasets, data that have high semantic correlations with in-distribution data are excluded from the OOD data for testing purposes.

\noindent \textbf {MOOD}: 
MOOD comprises 11 publicly available datasets: Texture, SVHN, CIFAR, Places365, LSUN-Crop, LSUN-Resize, iSUN, MNIST, KMNIST, Fashion MNIST, and STL. In comparison to SC-OOD, MOOD encompasses a more diverse range of OOD test data.

\noindent \textbf {SYN}:  
SYN consists of 3 synthetic noise-based OOD test datasets, specifically Gaussian noise ($\sigma$=0.5), Rademacher Noise, and Blob. To control the randomness of the noise, we utilize random noise data with a seed fixed at 1.

\section*{ \large Implementation Details}
\subsection{Training Details}
\label{train_detail}
We employ the ResNet-18 model for our experiments. For training a model from scratch, we adopt the training settings from PASCL~\cite{wang2022partial} with the data augmentation used in OE~\cite{hendrycks2018deep}. Specifically, we train the model for 200 epochs using a learning rate of 0.001, weight decay of 0.0005, batch size of 128, cosine annealing scheduler, and Adam optimizer. To address the instability issue when training with long-tailed in-distribution datasets in the OE setting, we adopt the training parameter settings from PASCL, with the exception of data augmentation.

For fine-tuning the pre-trained model, we follow the original setting of OE. We apply the following settings: a learning rate of 0.001, weight decay of 0.0005, an in-distribution batch size of 128, an auxiliary batch size of 256, cosine annealing scheduler, and SGD optimizer.
When the in-distribution dataset is CIFAR~\cite{krizhevsky2009learning}, we fine-tune the model for 10 epochs. However, when dealing with the LT-CIFAR~\cite{cao2019learning} dataset, which is a long-tailed in-distribution dataset, we fine-tune the model for 20 epochs. 
This longer fine-tuning duration of 20 epochs is employed because the classification task for long-tailed datasets is consuming more time to converge compared to balanced datasets.
\subsection{Hyperparameter Tunings}
\label{hyperparameter_tune}
In our work, we have newly proposed hyperparameters in the following sections: 1) Self Knowledge Distillation 
2) Semi-hard Outlier Sampling
3) Outlier-aware Supervised Contrastive Learning.

In Section 1, we handle $\lambda_{KD}$ and $T_{KD}$ similarly to the original work~\cite{hinton2015distilling}, setting $\lambda_{KD}=1$ and searching for the optimal value of $T_{KD}$ within the set $\{1,2,4,5,10\}$. The highest classification accuracy is achieved with $T_{KD}=4$, so we fix $T_{KD}=4$ for all tasks. Most algorithms perform well with $\lambda_{KD}=1.0$. However, exceptionally in EnergyOE, setting $\lambda_{KD}=1.0$ improves accuracy but degrades OOD detection performance, hence we lower $\lambda_{KD}=0.1$ in this case.

In Section 2, we handle $q$, $step$. We select the value of $q \in \{0.00, 0.05, 0.10, \dots, 0.90\}$ and the value of $step \in \{1, 2\}$ that yield the best OOD detection performance ( $q \in \{0.00, 0.05, 0.10, \dots, 0.80\}$ for the case of $step=2$).

In Section 3, we introduce $\lambda_{SC}$, $\tau_{sc}$, and $n$. We adopt $\tau_{sc}=0.1$ following the original work~\cite{khosla2020supervised}. During fine-tuning, we identify the optimal $\lambda_{SC} \in \{0.1, 0.2, 0.5, 1.0, 2.0, 5.0\}$ that yields the best OOD detection performance. Ultimately, we set $\lambda_{SC}=1$. As indicated in Table~\ref{C_ablation_batch}, we experiment with $n$ values from the set $\{2, 4, 6, 8\}$. However, for simplicity, we fix $n=4$ for all tasks, given its effective performance and computational efficiency.

\subsection{Implementation Details for The Proposed Method}
We commonly follow the approach of training a base model from scratch, as outlined in Training Details Section, and then using this pre-trained model for fine-tuning.
We provide implementation details for applying our algorithm to OECC~\cite{papadopoulos2021outlier}, EnergyOE~\cite{liu2020energy}, and Balanced EnergyOE~\cite{choi2023}. We mostly follow the hyperparameter setting of original works. Into our new parameters, we fix $\lambda_{SC}=1$, $\tau_{sc}=0.1$, $T_{KD}=4$, and $n=4$ for the $n$-batch transform across all algorithms.

For OECC, we assign the coefficients of regularization losses ($\lambda_{reg}$) as follows: $\lambda_1=0.07$ and $\lambda_2=0.05$. However, for LT-CIFAR datasets, we lower the coefficients to $\lambda_1 = \lambda_2 = 0.03$. This adjustment aims to prevent training instability.
Regarding knowledge distillation, we set $\lambda_{KD}=1$ for all in-distribution (ID) datasets.
For outlier sampling, we employ specific quantile $q$ values. Specifically, we use 0.7 (with step-size=2) for CIFAR10, 0.85 (with step-size=1) for CIFAR100, 0.8 (with step-size=2) for LT-CIFAR10, and 0.9 (with step-size=1) for LT-CIFAR100.


Moving on to EnergyOE, we utilize specific parameter settings for different datasets. For CIFAR datasets, we set $\lambda_{reg}=0.1$ and $\lambda_{KD}=0.1$. However, for LT-CIFAR datasets, we adjust the values to $\lambda_{reg}=0.15$ and $\lambda_{KD}=1$.
To incorporate the proposed outlier sampling method into EnergyOE, we use different quantile $q$ values. Specifically, we choose 0.7 (with step-size=2) for CIFAR10, no sampling for CIFAR100, 0.75 (with step-size=2) for LT-CIFAR10, and 0.9 (with step-size=1) for LT-CIFAR100. We maintain the same hyperparameter values ($m_{in}$ and $m_{out}$) as those used in the EnergyOE.

For the Balanced EnergyOE, we adopt a consistent parameter setting across LT-CIFAR10 and LT-CIFAR100. We set $\lambda_{reg}=0.15$ and $\lambda_{KD}=1$ for regularization and knowledge distillation, respectively. Additionally, the newly introduced parameters $\gamma$ and $\alpha$ align with the Balanced EnergyOE setting.
Specifically, $\gamma=0.75$ for all datasets. As for $\alpha$, we use a value of 10 for LT-CIFAR10 and 100 for LT-CIFAR100. 
Furthermore, we maintain the same hyperparameter values ($m_{in}$ and $m_{out}$) as those employed in the Balanced EnergyOE (Same value as the EnergyOE).
In certain cases, we have observed that incorporating outlier sampling into the EnergyOE and Balanced EnergyOE methods results in a reduction in performance on the SC-OOD benchmark, even though it yields improvements on the MOOD benchmark. Therefore, we do not employ outlier sampling for these specific cases.
\subsection{Implementation Details for Other Methods}
For MSP~\cite{hendrycks2016baseline}(ST) and Energy~\cite{liu2020energy}(ST), we obtain the score from a base model. The training procedure for the base model is described in Training Details Section.

For OE~\cite{hendrycks2018deep}(scratch) training from a scratch model, we follow the original setting $\lambda_{reg}=0.5$ of OE. The training procedure for the scratch model is described in Training Details Section.

For EnergyOE~\cite{liu2020energy}(scratch) training from a scratch model, we use $\lambda_{reg}=0.01$ which is different from the original setting for fine-tuning $\lambda_{reg}=0.1$ of EnergyOE. Because when we set $\lambda_{reg}=0.1$, classification accuracy is degraded to 50\% for the CIFAR100 dataset (When $\lambda_{reg}=0.01$, classification accuracy is 65.92\%)

For the implementation of PASCL\footnote{\url{https://github.com/amazon-science/long-tailed-ood-detection}}~\cite{wang2022partial}, OS\footnote{\url{https://github.com/hongxin001/open-sampling}}~\cite{wei2022open}, WOOD\footnote{\url{https://github.com/jkatzsam/woods_ood}}~\cite{katz2022training}, and NTOM\footnote{\url{https://github.com/jfc43/informative-outlier-mining}}~\cite{chen2021atom}, we mostly follow the publicly available code.
There are three main differences in our implementation compared to the original code. Firstly, we use the same auxiliary data setting as PASCL, where the auxiliary data is taken from the first 30K images of the 300K Random Images dataset~\cite{hendrycks2018deep}. Secondly, we unify the neural network model to ResNet-18 for consistency across the experiments. Thirdly, we consistently obtain eight models by fixing all random seeds from 1 to 8 for each method and calculating the average and standard deviation of the results.

In the implementation of NTOM, we adjust the data sampling pool size to align with our auxiliary data set, which differs from the original setting. NTOM originally set $N=400,000$, $n=100,000$ for sampling. However, since our auxiliary dataset contains 300K data, we set $N=200,000$, $n=100,000$ for NTOM (100\%) and $N=30,000$, $n=15,000$ for NTOM (10\%).

For the implementation of ReAct\footnote{\url{https://github.com/deeplearning-wisc/react}}~\cite{sun2021react} and kNN\footnote{\url{https://github.com/deeplearning-wisc/knn-ood}}~\cite{sun2022out}, we mostly follow the pubilicly available code. We use these post-hoc scoring methods based on our base pre-trained model. For the implementation of ReAct, we use the best threshold in set $\{1, 2, 3, 4, 5\}$ which gives the best OOD detection performance (AUROC). For the implementation of kNN, we use the best nearest neighbor number $k$ in set $\{1, 2, 5, 10, 25, 50\}$ which gives the best OOD detection performance (AUROC).

\subsection{Implementation Details for Large-scale Imbalanced dataset}

The ImageNet-LT experiment mostly follows the same OE setting as the CIFAR experiment. In ImageNet-LT, the original protocol in PASCL~\cite{wang2022partial} uses all auxiliary data without sub-sampling. For a fair comparison, we excluded the Semi-hard data sampling methodology and applied the remaining SKD and OSCL methods. Hyperparameters were set as follows: $\lambda_{KD}=1$, $\lambda_{SC}=1$, $T_{KD}=20$, and $n=4$. Interestingly, increasing $T_{KD}$ in  SKD is beneficial in improving accuracy. As shown in Table~\ref{Image_ablation}, a slight increase in $\lambda_{reg}$ leads to a substantial drop in accuracy for ImageNet-LT. To achieve high accuracy and excellent OOD detection performance, we set $\lambda_{reg}=1.25$.

\begin{table}[h!]
\caption{Effect of regularization coefficient $\lambda_{reg}$ on Large-scale dataset. Result on LT-ImageNet benchmark (mean over 5 datasets).}
\centering
\scriptsize
\label{Image_ablation}
\begin{tabular}{|c|c|c|ccc|}
\hline
\multirow{2}{*}{METHOD}                                             & \multirow{2}{*}{$\lambda_{reg}$} & \multirow{2}{*}{ACC$\uparrow$} & \multicolumn{3}{c|}{OOD detection}                                                         \\ \cline{4-6} 
                                                                    &                         &                      & \multicolumn{1}{c|}{FPR$\downarrow$}            & \multicolumn{1}{c|}{AUROC$\uparrow$}          & AUPR$\uparrow$           \\ \hline
OE                                                                  & 1.0                     & 38.46                & \multicolumn{1}{c|}{73.45}          & \multicolumn{1}{c|}{75.07}          & 47.72          \\ \hline
\multirow{9}{*}{\begin{tabular}[c]{@{}c@{}}OE\\ +ours\end{tabular}} & 1.0                     & \textbf{39.36}       & \multicolumn{1}{c|}{70.67}          & \multicolumn{1}{c|}{78.99}          & 54.83          \\
                                                                    & \textbf{1.25}           & 39.20                & \multicolumn{1}{c|}{69.58}          & \multicolumn{1}{c|}{79.51}          & 54.96          \\
                                                                    & 1.50                    & 38.63                & \multicolumn{1}{c|}{69.90}          & \multicolumn{1}{c|}{79.48}          & 54.65          \\
                                                                    & 1.75                    & 37.17                & \multicolumn{1}{c|}{70.48}          & \multicolumn{1}{c|}{78.77}          & 53.46          \\
                                                                    & 2.00                    & 38.50                & \multicolumn{1}{c|}{69.29}          & \multicolumn{1}{c|}{79.87}          & \textbf{55.41} \\
                                                                    & 2.25                    & 38.21                & \multicolumn{1}{c|}{69.84}          & \multicolumn{1}{c|}{79.16}          & 53.42          \\
                                                                    & 2.50                    & 38.18                & \multicolumn{1}{c|}{\textbf{68.16}} & \multicolumn{1}{c|}{\textbf{80.04}} & 54.89          \\
                                                                    & 2.75                    & 38.08                & \multicolumn{1}{c|}{68.25}          & \multicolumn{1}{c|}{80.03}          & 55.20          \\
                                                                    & 3.00                    & 37.91                & \multicolumn{1}{c|}{68.87}          & \multicolumn{1}{c|}{79.83}          & 54.69          \\ \hline
\end{tabular}
\end{table}

\section*{ \large Visualization}



\subsection{Visualization of Hardness}


\begin{figure*}[t!]
\centering

\begin{subfigure}[b]{0.3\linewidth}
        \caption{}
        \includegraphics[width=\linewidth]{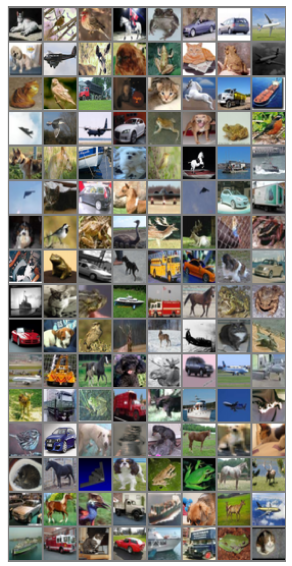}
        \label{fig_id_cifar10}
\end{subfigure}
\begin{subfigure}[b]{0.3\linewidth}
        \caption{}
        \includegraphics[width=\linewidth]{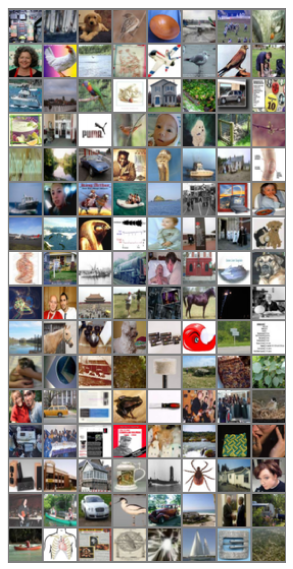}
        \label{fig_ood_hard}
\end{subfigure}
\begin{subfigure}[b]{0.3\linewidth}
        \caption{}
        \includegraphics[width=\linewidth]{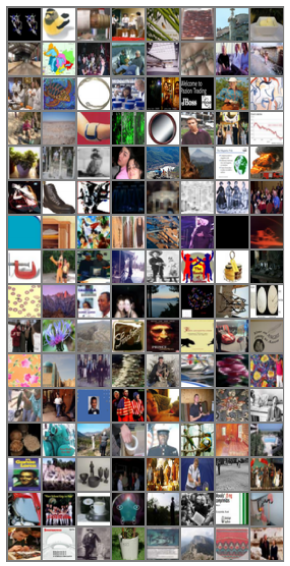}
        \label{fig_ood_easy}
\end{subfigure}

\caption{\textbf{Qualitative results for showing the effect of outlier samples'  hardness when ID dataset is CIFAR10.} Figure~\ref{fig_id_cifar10} shows in-distribution training data.  Figure~\ref{fig_ood_hard} presents outlier hard samples whose hardness quantile $q$ is between 0.9 to 1.0. Figure~\ref{fig_ood_easy} presents outlier easy samples whose hardness quantile $q$ is between 0.0 to 0.1. The sampling step size $step$ is set to 1.
}
\label{fig_visualize_hardness}

\end{figure*}

We qualitatively validate the effectiveness of our defined hardness by examining the CIFAR10 dataset. The training samples for CIFAR10 are represented in (a). Based on the hardness criteria, hard samples and easy samples are identified and represented in (b) and (c), respectively.
Through visualization, we can observe that hard samples exhibit a strong semantic correlation with the CIFAR10 training samples. On the other hand, easy samples have a lower semantic correlation. 

In particular, extremely hard samples often exhibit a resemblance to actual in-distribution samples, as illustrated in (b).
Consequently, attempting to distinguish these samples as outliers from the in-distribution becomes a challenging task.
As a result, this task makes training unstable which makes both OOD performance and classification accuracy deteriorate, as depicted in Figure~\ref{fig_num_quantile} in the main text.
This visual confirmation strengthens the academic validity of our findings.

\section*{ \large Detailed Experimental Results}

In our validation process, we expand the evaluation by including both the MOOD benchmark and the SYN (synthetic) benchmark. The SYN dataset is proposed as an out-of-distribution test dataset in the OE~\cite{hendrycks2018deep}. It comprises three types of synthetic data: Gaussian, Rademacher, and Blobs. To ensure consistency, we fix the random seed to 1, maintaining a consistent set of random synthetic data throughout our experiments on the SYN dataset. This approach establishes a standardized and reproducible evaluation framework for assessing the performance of our method on synthetic data.

In the Expanded Result Section, we provide a concise presentation of the detailed dataset-specific results for each benchmark, focusing on the comparison between OE and our proposed method. In the Detailed SC-OOD Result Section, we present comparison results for the SC-OOD benchmark, including the standard deviation that couldn't be included in the main paper due to space constraints. We observe that fine-tuning-based methods exhibit a lower standard deviation compared to methods that are from scratch, indicating more consistent performance across different random seeds.
In the Detailed MOOD Result Section, we summarize the comparison results for the MOOD benchmark, including the standard deviation, as previously mentioned. Similarly, in the Detailed SYN Result Section, we provide a summary of the comparison results for the SYN benchmark, incorporating the standard deviation, as discussed earlier.

\subsection{Expanded Result on each OOD dataset}
\label{expanded_result}
We summarize the results of comparing OE and our method for each dataset. Overall, our proposed method consistently outperforms OE in terms of OOD detection performance across the majority of datasets. This improvement in performance is observed not only in the MOOD benchmark but also in the SYN benchmark. These findings highlight the effectiveness of our method in enhancing OOD detection capabilities compared to the baseline OE approach.
\newpage
{\bf\noindent Expanded result on SC-OOD:}
\begin{table*}[h!]
\caption{Expanded result on SC-OOD benchmark by each dataset.}
\label{SC_OOD_dataset}
\centering
\scriptsize
\begin{tabular}{cc|cc|cc|cc}
\hline
\multicolumn{2}{c|}{Dataset} & \multicolumn{2}{c|}{FPR $\downarrow$} & \multicolumn{2}{c|}{AUROC$\uparrow$} & \multicolumn{2}{c}{AUPR$\uparrow$} \\ \hline
\multicolumn{1}{c|}{$D_{in}$}                        & $D_{out}^{test}$                                                   & OE & OE+ours & OE & OE+ours & OE & OE+ours \\ \hline
\multicolumn{1}{c|}{\multirow{7}{*}{CIFAR10}}     & Texture                                                  & 12.50 $\pm 0.23$   &   6.03 $\pm 0.39$      &  97.66 $\pm 0.04$  &   98.70 $\pm 0.08$      & 95.94 $\pm 0.08$   &   97.48 $\pm 0.19$      \\
\multicolumn{1}{c|}{}                             & SVHN                                                     & 6.62 $\pm 0.25$   &   4.50 $\pm 1.05$      & 98.67 $\pm 0.03$   &  99.00 $\pm 0.24$       & 99.42 $\pm 0.02$   &  99.50 $\pm 0.11$       \\
\multicolumn{1}{c|}{}                             & CIFAR100                                                 &  31.62 $\pm 0.20$  &  28.49 $\pm 0.40$       &  91.76 $\pm 0.03$  &   93.16 $\pm 0.09$      & 90.92 $\pm 0.03$   &   92.65 $\pm 0.09$      \\
\multicolumn{1}{c|}{}                             & Tiny ImageNet &  25.98 $\pm 0.12$  &  21.97 $\pm 0.40$       & 92.82 $\pm 0.03$   &  94.53 $\pm 0.07$       &  89.78 $\pm 0.04$  &  91.98 $\pm 0.15$       \\
\multicolumn{1}{c|}{}                             & LSUN                                                     & 16.43 $\pm 0.12$   &  9.02 $\pm 0.53$       & 96.27 $\pm 0.03$   &  97.89 $\pm 0.08$       & 95.77 $\pm 0.03$   &  97.48 $\pm 0.06$       \\
\multicolumn{1}{c|}{}                             & Places365                                                &   19.75 $\pm 0.15$ &   12.88 $\pm 0.29$      & 95.28 $\pm 0.03$   &  96.95 $\pm 0.06$       &  98.14 $\pm 0.01$  & 98.77 $\pm 0.03$        \\ \cline{2-8} 
\multicolumn{1}{c|}{}                             & Average                                                &  18.82 $\pm 0.07$  &    \textbf{13.81} $\pm 0.22$     &  95.41 $\pm 0.02$  &  \textbf{96.70} $\pm 0.06$       & 94.99 $\pm 0.02$   & \textbf{96.31} $\pm 0.08$        \\ \hline

\multicolumn{1}{c|}{\multirow{7}{*}{CIFAR100}}    & Texture                                                  & 51.82 $\pm 0.53$   &  43.24 $\pm 1.20$       & 83.02 $\pm 0.22$   &   87.86 $\pm 0.30$      & 70.31 $\pm 0.33$   &   78.90 $\pm 0.51$      \\
\multicolumn{1}{c|}{}                             & SVHN                                                     & 49.91 $\pm 1.66$   &   22.55 $\pm 1.79$      & 83.24 $\pm 0.77$   & 94.08 $\pm 0.62$        & 90.17 $\pm 0.48$   & 96.97 $\pm 0.38$        \\
\multicolumn{1}{c|}{}                             & CIFAR10                                                  & 64.16 $\pm 0.27$   &  59.87 $\pm 0.70$       & 75.00 $\pm 0.11$   &   75.94 $\pm 0.41$      &  69.70 $\pm 0.14$  &   70.50 $\pm 0.64$      \\
\multicolumn{1}{c|}{}                             & Tiny ImageNet & 58.10 $\pm 0.17$   &   56.46 $\pm 0.41$      &  79.03 $\pm 0.05$  &   79.05 $\pm 0.19$      &  64.58 $\pm 0.09$  &   64.57 $\pm 0.27$      \\
\multicolumn{1}{c|}{}                             & LSUN                                                     & 48.00 $\pm 0.32$   &  37.48 $\pm 0.77$       & 85.49 $\pm 0.11$   &  87.98 $\pm 0.28$       &  74.18 $\pm 0.16$  & 77.49 $\pm 0.51$        \\
\multicolumn{1}{c|}{}                             & Places365                                                & 48.32 $\pm 0.30$   &  39.38 $\pm 0.37$       & 85.94 $\pm 0.10$   &  88.64 $\pm 0.21$       & 92.79 $\pm 0.07$   & 94.20 $\pm 0.14$        \\ \cline{2-8} 
\multicolumn{1}{c|}{}                             & Average                                      &   53.38 $\pm 0.33$ &    \textbf{43.16} $\pm 0.49$     & 81.96 $\pm 0.12$   &   \textbf{85.59} $\pm 0.22$      &  76.95 $\pm 0.11$  &  \textbf{80.44} $\pm 0.26$       \\ \hline

\multicolumn{1}{c|}{\multirow{7}{*}{LT-CIFAR10}}  & Texture                                                  & 45.45 $\pm 0.07$   &  27.22 $\pm 0.37$       & 87.87 $\pm 0.08$   &   93.89 $\pm 0.08$      & 79.85 $\pm 0.19$   &   89.60 $\pm 0.18$      \\
\multicolumn{1}{c|}{}                             & SVHN                                                     & 27.44 $\pm 0.44$   &  24.97 $\pm 1.15$       & 92.21 $\pm 0.11$   &  94.31 $\pm 0.32$       &  95.58 $\pm 0.07$  &  97.31 $\pm 0.17$       \\
\multicolumn{1}{c|}{}                             & CIFAR100                                                 & 64.88 $\pm 0.17$   &   56.97 $\pm 0.24$      &  78.35 $\pm 0.07$  &  84.48 $\pm 0.08$       & 76.47 $\pm 0.07$   &   83.87 $\pm 0.09$      \\
\multicolumn{1}{c|}{}                             & Tiny ImageNet & 58.62 $\pm 0.30$   &   47.85 $\pm 0.40$      &  81.51 $\pm 0.07$  &   87.22 $\pm 0.04$      &  75.81 $\pm 0.08$  &  82.77 $\pm 0.07$       \\
\multicolumn{1}{c|}{}                             & LSUN                                                     &  54.08 $\pm 0.50$  &   29.80 $\pm 0.38$      & 86.32 $\pm 0.18$   &   93.33 $\pm 0.09$      &  85.97 $\pm 0.13$  &   92.75 $\pm 0.10$      \\
\multicolumn{1}{c|}{}                             & Places365                                                &  58.60 $\pm 0.45$  &  36.33 $\pm 0.38$       & 84.39 $\pm 0.16$   &   91.59 $\pm 0.07$      &  93.88 $\pm 0.05$  &   96.69 $\pm 0.03$      \\ \cline{2-8} 
\multicolumn{1}{c|}{}                             & Average                                                & 51.51 $\pm 0.15$   &  \textbf{37.19} $\pm 0.27$       & 85.11 $\pm 0.06$   &   \textbf{90.81} $\pm 0.08$      & 84.59 $\pm 0.05$   &    \textbf{90.50} $\pm 0.07$     \\ \hline

\multicolumn{1}{c|}{\multirow{7}{*}{LT-CIFAR100}} & Texture                                                  &  84.16 $\pm 0.18$  &  75.60 $\pm 0.61$       &  66.17 $\pm 0.13$  &  78.76 $\pm 0.17$       &  51.85 $\pm 0.17$  &   70.27 $\pm 0.26$      \\
\multicolumn{1}{c|}{}                             & SVHN                                                     &  64.64 $\pm 0.27$  &   50.39 $\pm 1.22$      & 74.70 $\pm 0.16$   &  81.77 $\pm 0.77$       & 85.33 $\pm 0.10$   &  89.07 $\pm 0.55$       \\
\multicolumn{1}{c|}{}                             & CIFAR10                                                  & 84.80 $\pm 0.12$   &    82.07 $\pm 0.26$     &  59.51 $\pm 0.05$  &  60.22 $\pm 0.16$       & 56.43 $\pm 0.06$   & 55.45 $\pm 0.14$        \\
\multicolumn{1}{c|}{}                             & Tiny ImageNet & 79.96 $\pm 0.15$   &  77.04 $\pm 0.15$       &  66.21 $\pm 0.04$  &   68.94 $\pm 0.07$      &  51.02 $\pm 0.04$  &  53.42 $\pm 0.10$       \\
\multicolumn{1}{c|}{}                             & LSUN                                                     & 72.96 $\pm 0.10$   &   58.68 $\pm 0.38$      & 73.38 $\pm 0.05$   &   81.08 $\pm 0.08$      &  59.00 $\pm 0.08$  &  68.23 $\pm 0.15$       \\
\multicolumn{1}{c|}{}                             & Places365                                                &  74.72 $\pm 0.10$  &   63.88 $\pm 0.15$      &  71.68 $\pm 0.04$  &   78.28 $\pm 0.04$      & 85.07 $\pm 0.03$   &  88.45 $\pm 0.03$       \\ \cline{2-8} 
\multicolumn{1}{c|}{}                             & Average                                   &  76.87 $\pm 0.07$  &    \textbf{67.95} $\pm 0.31$     &  68.61 $\pm 0.05$  &   \textbf{74.84} $\pm 0.15$      &  64.78 $\pm 0.04$  &   \textbf{70.81} $\pm 0.13$      \\ \hline

\end{tabular}
\end{table*}

\newpage
{\bf\noindent Expanded result on MOOD:}

\begin{table*}[h!]
\caption{Expanded result on MOOD benchmark by each dataset.}
\label{MOOD_dataset}
\centering
\scriptsize
\begin{tabular}{cc|cc|cc|cc}
\hline
\multicolumn{2}{c|}{Dataset}        & \multicolumn{2}{c|}{FPR $\downarrow$} & \multicolumn{2}{c|}{AUROC$\uparrow$} & \multicolumn{2}{c}{AUPR$\uparrow$} \\ \hline
\multicolumn{1}{c|}{$D_{in}$} &$D_{out}^{test}$  & OE        & OE+ours        & OE        & OE+ours        & OE       & OE+ours       \\ \hline
\multicolumn{1}{c|}{\multirow{13}{*}{CIFAR10}}     & CIFAR100      & 31.62 $\pm 0.20$ & 28.49 $\pm 0.40$ & 91.76 $\pm 0.03$ & 93.16 $\pm 0.09$ & 90.92 $\pm 0.03$ & 92.65 $\pm 0.09$ \\
\multicolumn{1}{c|}{}                              & MNIST         & 8.62 $\pm 0.54$ & 4.87 $\pm 0.87$ & 98.02 $\pm 0.12$ & 98.50 $\pm 0.26$ & 97.52 $\pm 0.16$ & 97.77 $\pm 0.38$ \\
\multicolumn{1}{c|}{}                              & K-MNIST       & 7.67 $\pm 0.31$ & 4.65 $\pm 0.59$ & 98.10 $\pm 0.08$ & 98.63 $\pm 0.17$ & 97.62 $\pm 0.11$ & 98.02 $\pm 0.28$ \\
\multicolumn{1}{c|}{}                              & fashion-MNIST & 10.72 $\pm 0.60$ & 6.40 $\pm 1.29$ & 97.97 $\pm 0.07$ & 98.38 $\pm 0.24$ & 97.87 $\pm 0.07$ & 97.99 $\pm 0.27$ \\
\multicolumn{1}{c|}{}                              & LSUN (crop)   & 1.90 $\pm 0.07$ & 1.79 $\pm 0.12$ & 99.63 $\pm 0.01$ & 99.59 $\pm 0.03$ & 99.62 $\pm 0.01$ & 99.49 $\pm 0.07$ \\
\multicolumn{1}{c|}{}                              & SVHN          & 5.03 $\pm 0.17$ & 3.53 $\pm 0.97$ & 98.92 $\pm 0.03$ & 99.20 $\pm 0.19$ & 98.80 $\pm 0.05$ & 98.97 $\pm 0.22$ \\
\multicolumn{1}{c|}{}                              & Textures      & 11.82 $\pm 0.24$ & 5.99 $\pm 0.33$ & 97.69 $\pm 0.04$ & 98.70 $\pm 0.08$ & 95.98 $\pm 0.07$ & 97.47 $\pm 0.19$ \\
\multicolumn{1}{c|}{}                              & STL10         & 89.40 $\pm 0.32$ & 85.96 $\pm 0.38$ & 68.32 $\pm 0.05$ & 68.90 $\pm 0.24$ & 63.29 $\pm 0.06$ & 63.28 $\pm 0.18$ \\
\multicolumn{1}{c|}{}                              & Places365     & 20.63 $\pm 0.21$ & 14.87 $\pm 0.47$ & 95.06 $\pm 0.05$ & 96.71 $\pm 0.06$ & 94.56 $\pm 0.06$ & 96.47 $\pm 0.08$ \\
\multicolumn{1}{c|}{}                              & iSUN          & 6.98 $\pm 0.31$ & 3.55 $\pm 0.44$ & 98.56 $\pm 0.07$ & 99.14 $\pm 0.10$ & 98.22 $\pm 0.09$ & 98.80 $\pm 0.15$ \\
\multicolumn{1}{c|}{}                              & LSUN (resize) & 6.46 $\pm 0.27$ & 3.33 $\pm 0.38$ & 98.65 $\pm 0.06$ & 99.18 $\pm 0.09$ & 98.47 $\pm 0.07$ & 98.95 $\pm 0.12$ \\ \cline{2-8} 
\multicolumn{1}{c|}{}                              & Average & 18.26 $\pm 0.11$ & \textbf{14.86} $\pm 0.23$ & 94.79 $\pm 0.11$ & \textbf{95.46} $\pm 0.04$ & 93.90 $\pm 0.04$ & \textbf{94.53} $\pm 0.09$ \\ \hline

\multicolumn{1}{c|}{\multirow{13}{*}{CIFAR100}}    & CIFAR10       & 64.16 $\pm 0.27$ & 59.87 $\pm 0.70$ & 75.00 $\pm 0.11$ & 75.94 $\pm 0.41$ & 69.70 $\pm 0.14$ & 70.50 $\pm 0.64$ \\
\multicolumn{1}{c|}{}                              & MNIST         & 11.34 $\pm 0.95$ & 15.51 $\pm 3.63$ & 97.33 $\pm 0.24$ & 95.61 $\pm 1.15$ & 96.86 $\pm 0.31$ & 93.87 $\pm 1.54$ \\
\multicolumn{1}{c|}{}                              & K-MNIST       & 15.48 $\pm 0.75$ & 11.00 $\pm 2.96$ & 95.93 $\pm 0.23$ & 96.85 $\pm 0.80$ & 94.82 $\pm 0.32$ & 95.56 $\pm 1.03$ \\
\multicolumn{1}{c|}{}                              & fashion-MNIST & 16.38 $\pm 0.52$ & 14.98 $\pm 1.03$ & 96.47 $\pm 0.12$ & 96.39 $\pm 0.30$ & 96.21 $\pm 0.14$ & 95.87 $\pm 0.35$ \\
\multicolumn{1}{c|}{}                              & LSUN (crop)   & 20.03 $\pm 0.45$ & 13.71 $\pm 0.53$ & 95.46 $\pm 0.15$ & 96.91 $\pm 0.16$ & 95.01 $\pm 0.21$ & 96.46 $\pm 0.20$ \\
\multicolumn{1}{c|}{}                              & SVHN          & 44.95 $\pm 1.45$ & 22.77 $\pm 1.75$ & 85.41 $\pm 0.70$ & 94.49 $\pm 0.67$ & 81.06 $\pm 0.92$ & 93.67 $\pm 0.91$ \\
\multicolumn{1}{c|}{}                              & Textures      & 52.71 $\pm 0.28$ & 45.19 $\pm 0.96$ & 82.89 $\pm 0.22$ & 87.65 $\pm 0.31$ & 70.31 $\pm 0.33$ & 78.95 $\pm 0.52$ \\
\multicolumn{1}{c|}{}                              & STL10         & 61.99 $\pm 0.29$ & 59.57 $\pm 0.48$ & 77.51 $\pm 0.11$ & 78.20 $\pm 0.43$ & 69.17 $\pm 0.17$ & 71.07 $\pm 0.84$ \\
\multicolumn{1}{c|}{}                              & Places365     & 65.14 $\pm 0.38$ & 58.58 $\pm 1.15$ & 77.16 $\pm 0.17$ & 82.67 $\pm 0.46$ & 73.32 $\pm 0.19$ & 79.46 $\pm 0.59$ \\
\multicolumn{1}{c|}{}                              & iSUN          & 63.60 $\pm 1.25$ & 52.78 $\pm 3.26$ & 79.12 $\pm 0.77$ & 84.71 $\pm 1.37$ & 74.65 $\pm 0.91$ & 80.54 $\pm 1.70$ \\
\multicolumn{1}{c|}{}                              & LSUN (resize) & 61.82 $\pm 1.19$ & 49.72 $\pm 3.35$ & 80.97 $\pm 0.75$ & 85.63 $\pm 1.38$ & 79.31 $\pm 0.82$ & 83.27 $\pm 1.61$ \\  \cline{2-8} 
\multicolumn{1}{c|}{}                              & Average & 43.42 $\pm 0.19$ & \textbf{36.70} $\pm 0.84$ & 85.75 $\pm 0.13$ & \textbf{88.64} $\pm 0.38$ & 81.86 $\pm 0.18$ & \textbf{85.38} $\pm 0.50$ \\ \hline

\multicolumn{1}{c|}{\multirow{13}{*}{LT-CIFAR10}}  & CIFAR100      & 64.88 $\pm 0.17$ & 56.97 $\pm 0.24$ & 78.35 $\pm 0.07$ & 84.48 $\pm 0.08$ & 76.47 $\pm 0.07$ & 83.87 $\pm 0.09$ \\
\multicolumn{1}{c|}{}                              & MNIST         & 29.57 $\pm 0.43$ & 13.01 $\pm 0.84$ & 93.45 $\pm 0.15$ & 97.01 $\pm 0.20$ & 93.16 $\pm 0.17$ & 96.47 $\pm 0.24$ \\
\multicolumn{1}{c|}{}                              & K-MNIST       & 31.87 $\pm 0.43$ & 14.80 $\pm 0.75$ & 92.51 $\pm 0.12$ & 96.53 $\pm 0.18$ & 92.06 $\pm 0.13$ & 96.05 $\pm 0.22$ \\
\multicolumn{1}{c|}{}                              & fashion-MNIST & 32.72 $\pm 0.40$ & 21.84 $\pm 0.59$ &  92.10 $\pm 0.12$& 94.73 $\pm 0.17$ & 91.81 $\pm 0.13$ & 94.13 $\pm 0.19$ \\
\multicolumn{1}{c|}{}                              & LSUN (crop)   & 15.97 $\pm 0.20$ & 17.21 $\pm 0.78$ & 95.94 $\pm 0.06$ & 97.00 $\pm 0.13$ & 95.40 $\pm 0.07$ & 96.99 $\pm 0.14$ \\
\multicolumn{1}{c|}{}                              & SVHN          & 25.53 $\pm 0.28$ & 28.91 $\pm 2.33$ & 91.92 $\pm 0.08$ & 93.49 $\pm 0.53$ & 88.62 $\pm 0.14$ & 92.77 $\pm 0.58$ \\
\multicolumn{1}{c|}{}                              & Textures      & 46.62 $\pm 0.30$ & 27.30 $\pm 0.21$ & 87.64 $\pm 0.08$ & 93.89 $\pm 0.07$ & 79.68 $\pm 0.19$ & 89.72 $\pm 0.18$ \\
\multicolumn{1}{c|}{}                              & STL10         & 91.94 $\pm 0.08$ & 91.36 $\pm 0.13$ & 59.12 $\pm 0.09$ & 60.55 $\pm 0.10$ & 54.71 $\pm 0.10$ & 55.66 $\pm 0.11$ \\
\multicolumn{1}{c|}{}                              & Places365     & 66.10 $\pm 0.49$ & 46.51 $\pm 0.55$ & 78.76 $\pm 0.23$ & 89.46 $\pm 0.10$ & 77.60 $\pm 0.21$ & 89.32 $\pm 0.12$ \\
\multicolumn{1}{c|}{}                              & iSUN          & 21.46 $\pm 0.27$ & 8.78 $\pm 0.27$ & 95.66 $\pm 0.06$ & 97.59 $\pm 0.08$ & 95.04 $\pm 0.08$ & 96.54 $\pm 0.20$ \\
\multicolumn{1}{c|}{}                              & LSUN (resize) & 21.18 $\pm 0.35$ & 8.35 $\pm 0.28$ & 95.84 $\pm 0.07$ & 97.80 $\pm 0.09$ & 95.73 $\pm 0.08$ & 97.15 $\pm 0.18$ \\  \cline{2-8} 
\multicolumn{1}{c|}{}                              & Average & 40.71 $\pm 0.20$ & \textbf{30.46} $\pm 0.43$ & 87.39 $\pm 0.07$ & \textbf{91.14} $\pm 0.09$ & 85.48 $\pm 0.08$ & \textbf{89.88} $\pm 0.11$ \\ \hline

\multicolumn{1}{c|}{\multirow{13}{*}{LT-CIFAR100}} & CIFAR10       & 84.80 $\pm 0.12$ & 82.07 $\pm 0.26$ & 59.51 $\pm 0.05$ & 60.22 $\pm 0.16$ & 56.43 $\pm 0.06$ & 55.45 $\pm 0.14$ \\
\multicolumn{1}{c|}{}                              & MNIST         & 45.62 $\pm 0.43$ & 14.81 $\pm 0.93$ & 84.29 $\pm 0.33$ & 94.13 $\pm 0.39$ & 79.11 $\pm 0.47$ & 88.88 $\pm 0.73$ \\
\multicolumn{1}{c|}{}                              & K-MNIST       & 51.52 $\pm 0.43$ & 16.37 $\pm 0.95$ & 82.42 $\pm 0.27$ & 94.56 $\pm 0.31$ & 77.36 $\pm 0.36$ & 90.72 $\pm 0.49$ \\
\multicolumn{1}{c|}{}                              & fashion-MNIST & 46.69 $\pm 0.75$ & 27.77 $\pm 0.71$ & 89.53 $\pm 0.11$ & 92.04 $\pm 0.13$ & 89.39 $\pm 0.10$ & 89.18 $\pm 0.18$ \\
\multicolumn{1}{c|}{}                              & LSUN (crop)   & 44.31 $\pm 0.21$ & 28.46 $\pm 0.29$ & 89.90 $\pm 0.05$ & 93.89 $\pm 0.07$ & 90.24 $\pm 0.06$ & 93.77 $\pm 0.10$ \\
\multicolumn{1}{c|}{}                              & SVHN          & 62.98 $\pm 0.24$ & 49.88 $\pm 1.41$ & 75.90 $\pm 0.16$ & 81.82 $\pm 0.79$ & 70.60 $\pm 0.18$ & 76.01 $\pm 1.05$ \\
\multicolumn{1}{c|}{}                              & Textures      & 84.42 $\pm 0.20$ & 76.43 $\pm 0.30$ & 66.26 $\pm 0.12$ & 78.50 $\pm 0.16$ & 52.09 $\pm 0.15$ & 70.01 $\pm 0.26$ \\
\multicolumn{1}{c|}{}                              & STL10         & 81.61 $\pm 0.11$ & 80.09 $\pm 0.33$ & 63.35 $\pm 0.07$ & 64.81 $\pm 0.17$ & 55.60 $\pm 0.08$ & 56.86 $\pm 0.21$ \\
\multicolumn{1}{c|}{}                              & Places365     & 78.32 $\pm 0.12$ & 64.61 $\pm 0.36$ & 68.09 $\pm 0.09$ & 77.60 $\pm 0.05$ & 64.92 $\pm 0.10$ & 73.55 $\pm 0.07$ \\
\multicolumn{1}{c|}{}                              & iSUN          & 75.07 $\pm 0.47$ & 60.61 $\pm 1.79$ & 68.38 $\pm 0.39$ & 78.83 $\pm 0.87$ & 60.77 $\pm 0.36$ & 72.33 $\pm 0.91$ \\
\multicolumn{1}{c|}{}                              & LSUN (resize) & 73.62 $\pm 0.57$ & 60.32 $\pm 1.52$ & 70.29 $\pm 0.35$ & 79.36 $\pm 0.82$ & 65.53 $\pm 0.35$ & 75.38 $\pm 0.83$ \\  \cline{2-8} 
\multicolumn{1}{c|}{}                              & Average & 66.27 $\pm 0.16$ & \textbf{51.04} $\pm 0.51$ & 74.36 $\pm 0.11$ & \textbf{81.43} $\pm 0.22$ & 69.28 $\pm 0.12$ & \textbf{76.56} $\pm 0.24$ \\ \hline

\end{tabular}
\end{table*}

\newpage
{\bf\noindent Expanded result on SYN:}

\begin{table*}[h!]
\caption{Expanded result on SYN benchmark by each dataset.}
\label{syn_dataset}
\centering
\scriptsize
\begin{tabular}{cc|cc|cc|cc}
\hline
\multicolumn{2}{c|}{Dataset} & \multicolumn{2}{c|}{FPR $\downarrow$} & \multicolumn{2}{c|}{AUROC$\uparrow$} & \multicolumn{2}{c}{AUPR$\uparrow$} \\ \hline
\multicolumn{1}{c|}{$D_{in}$}                        & $D_{out}^{test}$      & OE & OE+ours & OE & OE+ours & OE & OE+ours \\ \hline
\multicolumn{1}{c|}{\multirow{4}{*}{CIFAR10}}     & Gaussian   & 0.96 $\pm 0.07$   &  1.66 $\pm 0.19$       & 99.52 $\pm 0.04$   &  98.87 $\pm 0.17$       &  98.59 $\pm 0.17$  &   96.00 $\pm 0.54$      \\
\multicolumn{1}{c|}{}                             & Rademacher &  1.15 $\pm 0.07$  & 2.24 $\pm 0.24$        & 99.41 $\pm 0.05$   &  98.28 $\pm 0.25$       & 98.24 $\pm 0.19$   &    93.84 $\pm 0.85$     \\
\multicolumn{1}{c|}{}                             & Blobs      &  9.79 $\pm 0.41$  &  2.62 $\pm 0.64$       &  97.39 $\pm 0.13$  &   99.26 $\pm 0.24$      & 96.06 $\pm 0.24$   &    98.88 $\pm 0.43$     \\ \cline{2-8} 
\multicolumn{1}{c|}{}                             & Average    & 3.97 $\pm 0.14$   &  \textbf{2.18} $\pm 0.29$       &  98.77 $\pm 0.05$  &    \textbf{98.80} $\pm 0.19$     & \textbf{97.63} $\pm 0.13$   &    96.24 $\pm 0.54$     \\ \hline
\multicolumn{1}{c|}{\multirow{4}{*}{CIFAR100}}    & Gaussian   & 29.06 $\pm 1.92$   &  5.59 $\pm 0.98$       & 78.72 $\pm 1.53$   &      97.55 $\pm 0.52$   &  63.62 $\pm 1.47$  &    94.21 $\pm 1.26$     \\
\multicolumn{1}{c|}{}                             & Rademacher &  0.75 $\pm 0.21$  &  1.31 $\pm 0.14$       & 99.70 $\pm 0.10$   &   99.06 $\pm 0.10$      & 99.25 $\pm 0.27$   &   96.39 $\pm 0.38$      \\
\multicolumn{1}{c|}{}                             & Blobs      &  9.27 $\pm 0.46$  &  4.43 $\pm 0.88$       &  97.41 $\pm 0.16$  &    99.05 $\pm 0.20$     &  96.55 $\pm 0.24$  &   98.86 $\pm 0.24$      \\ \cline{2-8} 
\multicolumn{1}{c|}{}                             & Average    &  13.03 $\pm 0.66$  & \textbf{3.78} $\pm 0.49$        &  91.94 $\pm 0.52$  &   \textbf{98.55} $\pm 0.19$      & 86.48 $\pm 0.53$   &   \textbf{96.48} $\pm 0.43$      \\ \hline
\multicolumn{1}{c|}{\multirow{4}{*}{LT-CIFAR10}}  & Gaussian   &  5.36 $\pm 0.29$  &   2.32 $\pm 0.25$      &  96.96 $\pm 0.17$  &   98.71 $\pm 0.18$      & 91.70 $\pm 0.44$   &   96.09 $\pm 0.59$      \\
\multicolumn{1}{c|}{}                             & Rademacher &  1.56 $\pm 0.06$  &  3.41 $\pm 0.24$       &  99.30 $\pm 0.04$  &   97.82 $\pm 0.18$      &  98.26 $\pm 0.11$  &  93.35 $\pm 0.57$       \\
\multicolumn{1}{c|}{}                             & Blobs      & 83.89 $\pm 0.53$   & 56.98 $\pm 1.13$        & 43.84 $\pm 1.17$   &   80.41 $\pm 0.72$      & 43.11 $\pm 0.65$   &   73.92 $\pm 0.84$      \\ \cline{2-8} 
\multicolumn{1}{c|}{}                             & Average    &  30.27 $\pm 0.10$  &  \textbf{20.90} $\pm 0.39$       & 80.03 $\pm 0.34$   &    \textbf{92.31} $\pm 0.27$     &  77.69 $\pm 0.12$  &   \textbf{87.79} $\pm 0.50$      \\ \hline
\multicolumn{1}{c|}{\multirow{4}{*}{LT-CIFAR100}} & Gaussian   & 69.33 $\pm 1.79$   &   20.93 $\pm 2.28$      & 56.99 $\pm 1.91$   &     90.32 $\pm 1.36$    & 48.77 $\pm 1.11$   &   81.02 $\pm 2.14$      \\
\multicolumn{1}{c|}{}                             & Rademacher & 74.69 $\pm 1.29$   &  38.85 $\pm 4.43$       & 49.13 $\pm 1.80$   &   77.21 $\pm 3.19$      & 44.56 $\pm 0.87$   &  63.66 $\pm 3.21$       \\
\multicolumn{1}{c|}{}                             & Blobs      &  74.48 $\pm 0.22$  &  26.83 $\pm 1.33$       & 69.74 $\pm 0.24$   &  91.67 $\pm 0.55$       &  65.63 $\pm 0.25$  &  88.25 $\pm 0.77$       \\ \cline{2-8} 
\multicolumn{1}{c|}{}                             & Average    & 72.83 $\pm 1.02$   &  \textbf{28.87} $\pm 2.02$       & 58.62 $\pm 1.22$   &    \textbf{86.40} $\pm 1.38$     & 52.99 $\pm 0.66$   &   \textbf{77.65} $\pm 1.58$      \\ \hline
\end{tabular}
\end{table*}

\subsection{Detailed Result on SC-OOD with Standard Deviation}
\label{detalied_result_SCOOD}


In the SC-OOD benchmark, when comparing the results of the existing method with our proposed method, we observe a significant difference in the average values of ACC, FPR, AUROC, and AUPR in relation to the standard deviation. This notable difference in the average values provides strong evidence for the superiority of our proposed method.
 
Furthermore, the standard deviation of our OOD detection results serves as additional evidence for the reliability of our approach. We can observe this by comparing the difference in the Z score. For example, the Z score of 2.807 corresponds to a 99.5\% confidence interval. The difference between OE and our method in Table~\ref{SCOOD_comparison_balanced_std} shows distinct variations in terms of FPR, AUROC, and AUPR for the CIFAR10 dataset. The differences are 5.64 (18.82-13.18) for FPR, 1.29 (96.70-95.41) for AUROC, and 1.31 (96.31-95.00) for AUPR. When these values are divided by their respective standard deviations (std) and converted into Z scores, they become 25.64, 21.50, and 16.38, respectively. This conversion to Z scores provides a measure of reliability for the significant differences.

The MSP(ST) and Energy(ST) models are pre-trained models that remain fixed throughout the experiments. Also, ReAct(ST) and kNN(ST) are post-scoring methods that remain fixed. As a result, they yield a standard deviation (std) of 0 since only a single run is performed. However, WOOD exhibits a unique behavior where it converges to the same OOD detection and accuracy regardless of the variation in random seed values. Consequently, the std value for WOOD also becomes 0.
\newpage
{\bf\noindent CIFAR model on SC-OOD:}
\begin{table}[h!]

\caption{ Comparison with other methods on CIFAR using ResNet18; Mean and std over 8 random runs are reported; SC-OOD benchmark detection average  (over 6 datasets) performance (FPR95, AUROC, AUPR) and classification accuracy (ACC) (a): Result on CIFAR10 (b): Result on CIFAR100.}
\begin{subtable}{1.0\linewidth}
\centering
\scriptsize
\caption{}
\label{balanced_cifar10_std}
\begin{tabular}{c|c|c|c|c}
\hline
Method         &ACC$\uparrow$     &FPR$\downarrow$     &      AUROC$\uparrow$   &     AUPR$\uparrow$         \\ \hline
MSP(ST)         & 93.69 $\pm 0.00$          & 31.32 $\pm 0.00$          & 89.25 $\pm 0.00$         & 86.63 $\pm 0.00$          \\
 Energy(ST)      & 93.69 $\pm 0.00$         & \textbf{29.07} $\pm 0.00$          & 91.55 $\pm 0.00$         & 89.88 $\pm 0.00$          \\
 ReAct(ST)~\cite{sun2021react}   & \textbf{93.70} $\pm 0.00$ & 29.18 $\pm 0.00$ & 91.56 $\pm 0.00$ & 89.90 $\pm 0.00$  \\ 
 kNN(ST)~\cite{sun2022out}   & 93.69 $\pm 0.00$ & 47.45 $\pm 0.00$ & \textbf{91.67} $\pm 0.00$ & \textbf{90.78} $\pm 0.00$  \\ \hline
  OE(scratch)       & 90.12 $\pm 0.69$         & 20.97 $\pm 1.83$         & 95.35 $\pm 0.41$        & 95.22 $\pm 0.38$         \\ 
   EnergyOE(scratch) & 90.50 $\pm 0.35$         & \textbf{17.91} $\pm 0.37$         & \textbf{95.85} $\pm 0.14$         & \textbf{95.78} $\pm 0.31$ \\
   WOOD(10\%)     & 93.18 $\pm 0.00$          &    21.72 $\pm 0.00$     &   94.23 $\pm 0.00$      &      92.94 $\pm 0.00$    \\   
   WOOD(100\%)     & 93.20 $\pm 0.00$         &    22.01 $\pm 0.00$     &   94.16 $\pm 0.00$      &      92.86 $\pm 0.00$    \\  
      NTOM(10\%)   & 84.37 $\pm 2.08$          &    47.22 $\pm 3.80$      &   82.63 $\pm 1.51$    &    79.97 $\pm 1.56$    \\
      NTOM(100\%)   & \textbf{94.39} $\pm 0.24$          &    22.24 $\pm 1.53$     &   92.49 $\pm 0.56$      & 90.61 $\pm 0.67$    \\ \hline 

    OE       & 93.12 $\pm 0.05$ & 18.82 $\pm 0.07$ & 95.41 $\pm 0.02$ & 95.00 $\pm 0.02$          \\ 
        \textbf{OE+Ours}           & \textbf{93.38} $\pm 0.10$ & \textbf{13.81} $\pm 0.22$ & \textbf{96.70} $\pm 0.06$ & \textbf{96.31} $\pm 0.08$          \\ \hline
        EnergyOE       & 93.33 $\pm 0.09$ & 14.45 $\pm 0.46$ & 96.81 $\pm 0.09$ & 96.73 $\pm 0.06$          \\ 
        \textbf{EnergyOE+Ours}           & \textbf{93.64} $\pm 0.06$ & \textbf{13.15} $\pm 0.27$ & \textbf{96.99} $\pm 0.06$ & \textbf{96.84} $\pm 0.06$          \\ \hline
        OECC        & 91.81 $\pm 0.24$ & 14.28 $\pm 0.23$ & 96.40 $\pm 0.15$ & 95.44 $\pm 0.44$          \\ 
        \textbf{OECC+Ours}           & \textbf{92.18} $\pm 0.28$ & \textbf{13.06} $\pm 0.52$ & \textbf{96.52} $\pm 0.22$ & \textbf{95.66} $\pm 0.43$         \\ \hline
       \end{tabular}
\end{subtable}
\begin{subtable}{1.0\linewidth}
\centering
\scriptsize
\caption{}
\begin{tabular}{c|c|c|c|c}
\hline
Method         &ACC$\uparrow$     &FPR$\downarrow$     &      AUROC$\uparrow$   &     AUPR$\uparrow$         \\ \hline
MSP(ST)         & \textbf{75.70} $\pm 0.00$         & 62.78 $\pm 0.00$          & 76.14 $\pm 0.00$        & 71.29 $\pm 0.00$         \\
 Energy(ST)      & \textbf{75.70} $\pm 0.00$         & 57.59 $\pm 0.00$          & 79.78 $\pm 0.00$         & 73.31 $\pm 0.00$         \\
 ReAct(ST)~\cite{sun2021react}   & 74.79 $\pm 0.00$ & \textbf{55.12} $\pm 0.00$ & \textbf{80.90} $\pm 0.00$ & 75.35 $\pm 0.00$  \\
 kNN(ST)~\cite{sun2022out}   & \textbf{75.70} $\pm 0.00$ & 76.65 $\pm 0.00$ & 79.13 $\pm 0.00$ & \textbf{76.86} $\pm 0.00$  \\ \hline
  OE(scratch)       & 66.47 $\pm 0.45$         & 61.50 $\pm 1.89$        & 80.71 $\pm 0.78$        & 77.44 $\pm 0.67$         \\ 
   EnergyOE(scratch) & 65.92  $\pm 0.77$        & 55.67 $\pm 1.99$         & 81.54 $\pm 0.54$         & 76.82 $\pm 0.47$ \\
   WOOD(10\%)     & 75.80 $\pm 0.00$          &     \textbf{51.09} $\pm 0.00$      &    \textbf{83.14} $\pm 0.00$      &       \textbf{77.83} $\pm 0.00$    \\   
   WOOD(100\%)     & \textbf{75.93} $\pm 0.00$          &    51.52 $\pm 0.00$     &   82.97 $\pm 0.00$      &      77.66 $\pm 0.00$    \\   
      NTOM(10\%)   & 49.81 $\pm 4.43$     &    81.31 $\pm 2.01$     &  63.57 $\pm 1.47$      &    60.03 $\pm 1.02$    \\ 
      NTOM(100\%)   & 72.41 $\pm 0.95$          &    61.77 $\pm 1.23$     &   75.58 $\pm 0.76$      &    70.71 $\pm 0.76$    \\ \hline

    OE       & \textbf{73.86} $\pm 0.10$ & 53.38 $\pm 0.33$ & 81.96 $\pm 0.12$ & 76.95 $\pm 0.11$          \\ 
        \textbf{OE+Ours}           & 73.13 $\pm 0.19$ & \textbf{43.16} $\pm 0.49$ & \textbf{85.59} $\pm 0.22$ & \textbf{80.44} $\pm 0.26$          \\ \hline
        EnergyOE       & 74.50 $\pm 0.16$ & 43.59 $\pm 0.98$ & 86.07 $\pm 0.26$ & 81.34 $\pm 0.19$          \\ 
        \textbf{EnergyOE+Ours}           & \textbf{74.93} $\pm 0.21$ & \textbf{42.46} $\pm 1.20$ & \textbf{86.38} $\pm 0.53$ & \textbf{81.49} $\pm 0.60$         \\ \hline
        OECC       & 69.36 $\pm 0.21$ & 45.41 $\pm 0.62$ & 84.01 $\pm 0.27$ & 77.84 $\pm 0.30$          \\ 
        \textbf{OECC+Ours}           & \textbf{72.23} $\pm 0.10$ & \textbf{41.78} $\pm 0.51$ & \textbf{85.64} $\pm 0.20$ & \textbf{80.26} $\pm 0.22$         \\ \hline
       \end{tabular}
\label{balanced_cifar100_std}
\end{subtable}
\label{SCOOD_comparison_balanced_std}
\end{table}

\newpage
{\bf\noindent LT-CIFAR model on SC-OOD:}
\begin{table}[h!]

\caption{ Comparison with other methods on LT-CIFAR using ResNet18; Mean and std over 8 random runs are reported; SC-OOD benchmark detection average (over 6 datasets) performance (FPR95, AUROC, AUPR) and classification accuracy (ACC) (a): Result on LT-CIFAR10 (b): Result on LT-CIFAR100.}
\begin{subtable}{1.0\linewidth}
\centering
\scriptsize
\caption{}
\label{imbalanced_cifar10_std}
\begin{tabular}{c|c|c|c|c}
\hline
Method         &ACC$\uparrow$     &FPR$\downarrow$     &      AUROC$\uparrow$   &     AUPR$\uparrow$         \\ \hline
MSP(ST)         & 73.28 $\pm 0.00$          & 61.30 $\pm 0.00$         & 74.55 $\pm 0.00$    & 72.26 $\pm 0.00$         \\
 Energy(ST)      & 73.28 $\pm 0.00$        & \textbf{53.82} $\pm 0.00$  & \textbf{80.33} $\pm 0.00$   & 77.02 $\pm 0.00$         \\
 ReAct(ST)~\cite{sun2021react}   & 73.28 $\pm 0.00$ & 53.83 $\pm 0.00$ & 80.33 $\pm 0.00$ & 77.02 $\pm 0.00$  \\
 kNN(ST)~\cite{sun2022out}  & 73.28 $\pm 0.00$ & 74.98 $\pm 0.00$ & 80.11 $\pm 0.00$ & \textbf{77.48} $\pm 0.00$  \\ \hline
  OE(scratch)       & 72.85 $\pm 1.26$         & 34.79 $\pm 0.63$         & 89.40 $\pm 0.38$        & 85.79 $\pm 0.68$         \\ 
   EnergyOE(scratch) & 73.40 $\pm 0.52$         & 34.36 $\pm 0.49$       & 86.52 $\pm 1.01$         & 81.56 $\pm 2.18$ \\
   PASCL~\cite{wang2022partial}     &  \textbf{77.08} $\pm 1.01$           &    33.60 $\pm 0.30$     &   \textbf{90.72} $\pm 0.28$      &    \textbf{88.89} $\pm 0.49$    \\    
      OS~\cite{wei2022open}  & 77.04 $\pm 0.54$          &   \textbf{30.74} $\pm 0.90$        &  90.33 $\pm 0.49$       & 85.61 $\pm 0.92$      \\ \hline   

    OE       & 69.73 $\pm 0.12$ & 51.51 $\pm 0.15$ & 85.11 $\pm 0.06$ & 84.59 $\pm 0.05$         \\ 
        \textbf{OE+Ours}           &  \textbf{77.67} $\pm 0.11$ &  \textbf{37.19} $\pm 0.27$ &  \textbf{90.81} $\pm 0.08$ &  \textbf{90.50} $\pm 0.07$          \\ \hline
        EnergyOE       & 74.77 $\pm 0.20$  & 33.74 $\pm 0.34$ & 91.86 $\pm 0.10$ & 91.91 $\pm 0.08$          \\ 
        \textbf{EnergyOE+Ours}           & \textbf{78.18} $\pm 0.13$ &  \textbf{32.15} $\pm 0.54$ & \textbf{92.10} $\pm 0.10$ &  \textbf{92.01} $\pm 0.08$          \\ \hline
        OECC       & 62.01 $\pm 0.26$ & 44.52 $\pm 0.20$ & 87.86 $\pm 0.05$ & 87.56 $\pm 0.07$          \\ 
        
        \textbf{OECC+Ours}           & \textbf{ 75.23} $\pm 0.16$ &  \textbf{32.86} $\pm 0.82$ &  \textbf{91.53} $\pm 0.19$ &  \textbf{90.49} $\pm 0.30$         \\\hline 
        BEnergyOE      & 76.30 $\pm 0.24$ & 30.94 $\pm 0.46$ & 92.52 $\pm 0.10$ &  \textbf{91.86} $\pm 0.09$          \\ 
        \textbf{BEnergyOE   +Ours}           &  \textbf{78.13} $\pm 0.18$ & \textbf{29.00} $\pm 0.17$ & \textbf{92.57} $\pm 0.06$ & 91.50 $\pm 0.08$         \\         
        \hline
       \end{tabular}
\end{subtable}
\begin{subtable}{1.0\linewidth}
\centering
\scriptsize
\caption{}
\begin{tabular}{c|c|c|c|c}
\hline
Method         &ACC$\uparrow$     &FPR$\downarrow$     &      AUROC$\uparrow$   &     AUPR$\uparrow$         \\ \hline
MSP(ST)         & \textbf{40.22} $\pm 0.00$         & 83.30 $\pm 0.00$        & 61.17 $\pm 0.00$    & 58.10 $\pm 0.00$         \\
 Energy(ST)      & \textbf{40.22} $\pm 0.00$         & 80.59 $\pm 0.00$     & 64.08 $\pm 0.00$       & 59.86 $\pm 0.00$         \\
 ReAct(ST)~\cite{sun2021react}   & 39.02 $\pm 0.00$ & \textbf{79.72} $\pm 0.00$ & \textbf{64.22} $\pm 0.00$ & 59.90 $\pm 0.00$  \\
 kNN(ST)~\cite{sun2022out}   & \textbf{40.22} $\pm 0.00$ & 87.97 $\pm 0.00$ & 62.91 $\pm 0.00$ & \textbf{60.41} $\pm 0.00$  \\ \hline
  OE(scratch)       & 41.31 $\pm 0.66$         & 69.93 $\pm 1.37$         & 73.32 $\pm 0.66$        & 67.92 $\pm 0.47$         \\ 
   EnergyOE(scratch) & 40.78 $\pm 0.33$         & 67.46 $\pm 1.81$       & 74.30 $\pm 0.64$         &  \textbf{70.09} $\pm 0.65$ \\
   PASCL~\cite{wang2022partial}     &  \textbf{43.10} $\pm 0.47$        &    67.51 $\pm 0.40$    &  73.40 $\pm 0.35$    & 67.02 $\pm 0.57$      \\   
      OS~\cite{wei2022open}   & 39.96 $\pm 0.60$          &     \textbf{66.84} $\pm 1.36$      &  \textbf{74.39} $\pm 0.71$      &  69.37 $\pm 0.58$    \\ \hline   

    OE~\cite{wei2022open}       & 38.93 $\pm 0.07$ & 76.87 $\pm 0.07$ & 68.61 $\pm 0.05$ & 64.78 $\pm 0.04$          \\ 
        \textbf{OE+Ours}           &  \textbf{39.73} $\pm 0.09$ &  \textbf{67.95} $\pm 0.31$ &  \textbf{74.84} $\pm 0.15$ &  \textbf{70.81} $\pm 0.14$          \\ \hline
        EnergyOE        & 40.54 $\pm 0.08$ & 64.54 $\pm 0.19$ & 76.33 $\pm 0.07$ & 72.13 $\pm 0.05$          \\ 
        
        \textbf{EnergyOE+Ours}           &  \textbf{40.76} $\pm 0.13$ & \textbf{60.34} $\pm 0.22$ & \textbf{ 77.50} $\pm 0.11$ &  \textbf{73.17} $\pm 0.09$         \\ \hline
        OECC        & 31.08 $\pm 0.11$ & 73.96 $\pm 0.41$ & 71.70 $\pm 0.15$ & 68.42 $\pm 0.09$          \\ 
        \textbf{OECC+Ours}           &  \textbf{39.77} $\pm 0.13$ &  \textbf{64.04} $\pm 0.25$ &  \textbf{75.91} $\pm 0.08$ &  \textbf{71.23} $\pm 0.07$         \\\hline 
        BEnergyOE    & 40.78 $\pm 0.21$ & 61.30 $\pm 0.26$ & 77.64 $\pm 0.07$ & 73.00 $\pm 0.04$          \\ 
        \textbf{BEnergyOE   +Ours}           & \textbf{43.45} $\pm 0.14$ &  \textbf{60.90} $\pm 0.38$ & \textbf{78.01} $\pm 0.10$ & \textbf{73.50} $\pm 0.10$         \\         
        \hline
       \end{tabular}
\label{imbalanced_cifar100_std}
\end{subtable}
\label{SCOOD_comparison_imbalanced_std}
\end{table}

\newpage

\subsection{Detailed Result on MOOD with Standard Deviation}
\label{detalied_result_MOOD}

In the MOOD benchmark, when comparing the results of the existing method with our proposed method, we observe a significant difference in the average values of ACC, FPR, AUROC, and AUPR in relation to the standard deviation. This substantial difference in the average values provides strong evidence for the superior performance of our proposed method.

{\bf\noindent CIFAR model on MOOD:}
\begin{table}[h!]
\caption{ Comparison result on MOOD benchmark using ResNet18; Mean and std over 8 random runs are reported; MOOD benchmark detection average (over 11 datasets) performance (FPR95, AUROC, AUPR) and classification accuracy (ACC) (a): Result on CIFAR10 (b): Result on CIFAR100.}
\begin{subtable}{1.0\linewidth}
\centering
\scriptsize
\caption{}
\label{MOOD_balanced_cifar10_std}
\begin{tabular}{c|c|c|c|c}
\hline

Method         &ACC$\uparrow$     &FPR$\downarrow$     &      AUROC$\uparrow$   &     AUPR$\uparrow$         \\ \hline
MSP(ST)         & 93.69 $\pm 0.00$         & 30.09 $\pm 0.00$     & 89.53 $\pm 0.00$        & 85.58 $\pm 0.00$         \\
 Energy(ST)      & 93.69 $\pm 0.00$         & \textbf{23.64} $\pm 0.00$         & \textbf{92.72}$\pm 0.00$       & \textbf{90.27} $\pm 0.00$         \\
 ReAct(ST)~\cite{sun2021react}   & \textbf{93.70} $\pm 0.00$ & 23.76 $\pm 0.00$ & 92.71 $\pm 0.00$ & 90.27 $\pm 0.00$  \\
 kNN(ST)~\cite{sun2022out}   & 93.69 $\pm 0.00$ & 37.03 $\pm 0.00$ & 91.78 $\pm 0.00$ & 91.88 $\pm 0.00$  \\
 \hline
  OE(scratch)       & 90.12 $\pm 0.69$         & 21.37 $\pm 2.59$         & 93.93 $\pm 0.56$        & 92.99 $\pm 0.66$       \\ 
   EnergyOE(scratch) & 90.50 $\pm 0.35$          & 19.24 $\pm 2.27$     & 94.08 $\pm 0.53$     & 92.91 $\pm 0.71$ \\
   
   WOOD(10\%)     & 93.18 $\pm 0.00$    &    16.59 $\pm 0.00$   &   95.43 $\pm 0.00$     &   \textbf{93.98} $\pm 0.00$    \\  
   
   WOOD(100\%)     & 93.20 $\pm 0.00$      &    \textbf{16.57} $\pm 0.00$  &   \textbf{95.44} $\pm 0.00$    &   93.94 $\pm 0.00$   \\ 
   
      NTOM(10\%)   & 84.37 $\pm 2.08$         &   46.13 $\pm 3.07$    &   82.37 $\pm 1.56$    &    76.72 $\pm 2.04$   \\
      
      NTOM(100\%)   & \textbf{94.39} $\pm 0.24$  &  23.88 $\pm 0.98$   &   92.23 $\pm 0.38$   &  89.44 $\pm 0.58$ \\\hline   

    OE      & 93.12 $\pm 0.05$ & 18.26 $\pm 0.11$ & 94.79 $\pm 0.03$ & 93.90 $\pm 0.04$          \\ 
    
        \textbf{OE+Ours}           & \textbf{93.38} $\pm 0.10$ & \textbf{14.86} $\pm 0.23$ & \textbf{95.46} $\pm 0.04$ & \textbf{94.53} $\pm 0.09$     \\ \hline
        
        EnergyOE       & 93.33 $\pm 0.09$ & 15.70 $\pm 0.27$ & 95.26 $\pm 0.08$ & 94.50 $\pm 0.12$          \\ 
        \textbf{EnergyOE+Ours}  & \textbf{93.64} $\pm 0.06$ & \textbf{14.40} $\pm 0.45$ & \textbf{95.53} $\pm 0.14$ & \textbf{94.77} $\pm 0.24$      \\ \hline
        
        OECC        & 91.81 $\pm 0.24$ & 15.42 $\pm 0.25$ & 94.95 $\pm 0.18$ & 93.27 $\pm 0.69$          \\ 
        
        \textbf{OECC+Ours}    & \textbf{92.18} $\pm 0.28$ & \textbf{14.60} $\pm 0.35$ & \textbf{95.23} $\pm 0.17$ & \textbf{93.66}  $\pm 0.50$   \\ \hline
       \end{tabular}
\end{subtable}
\begin{subtable}{1.0\linewidth}
\centering
\scriptsize
\caption{}
\begin{tabular}{c|c|c|c|c}
\hline

Method         &ACC$\uparrow$     &FPR$\downarrow$     &      AUROC$\uparrow$   &     AUPR$\uparrow$         \\ \hline
MSP(ST)         & \textbf{75.70} $\pm 0.00$         & 59.82 $\pm 0.00$   & 78.12 $\pm 0.00$     & 73.43 $\pm 0.00$         \\
 Energy(ST)      & \textbf{75.70} $\pm 0.00$     & 47.23 $\pm 0.00$    & 84.34 $\pm 0.00$  & 79.28 $\pm 0.00$    \\
 ReAct(ST)~\cite{sun2021react}   & 74.79 $\pm 0.00$ & \textbf{40.55} $\pm 0.00$ & \textbf{86.63} $\pm 0.00$ & \textbf{82.34} $\pm 0.00$  \\
 kNN(ST)~\cite{sun2022out}   & \textbf{75.70} $\pm 0.00$ & 72.69 $\pm 0.00$ & 79.90 $\pm 0.00$ & 79.46 $\pm 0.00$  \\
 \hline
  OE(scratch)       & 66.47 $\pm 0.45$         & 52.00 $\pm 3.94$    & 84.06 $\pm 1.67$    & 81.18 $\pm 1.88$     \\ 
   EnergyOE(scratch) & 65.92 $\pm 0.77$          & 42.67 $\pm 3.86$   & 86.16 $\pm 1.22$     & 81.84 $\pm 1.53$ \\
   
   WOOD(10\%)     & 75.80 $\pm 0.00$    &   \textbf{35.05} $\pm 0.00$    &   \textbf{89.03} $\pm 0.00$     &  \textbf{85.94} $\pm 0.00$   \\  
   
   WOOD(100\%)     & \textbf{75.93} $\pm 0.00$   &   35.66 $\pm 0.00$     &   88.82 $\pm 0.00$     &   85.69 $\pm 0.00$    \\ 
   
      NTOM(10\%)   & 49.81 $\pm 4.43$  &  80.57 $\pm 2.92$  & 61.35 $\pm 2.87$     &  56.30 $\pm 2.37$    \\
      
      NTOM(100\%)   &  72.41 $\pm 0.95$    &  52.33 $\pm 1.60$   &   80.98 $\pm 0.79$   & 76.65 $\pm 1.03$   \\\hline  

    OE       & \textbf{73.86} $\pm 0.10$  & 43.42 $\pm 0.20$ & 85.75 $\pm 0.13$ & 81.86 $\pm 0.18$          \\ 
    
        \textbf{OE+Ours}           & 73.13 $\pm 0.19$ & \textbf{36.70} $\pm 0.84$ & \textbf{88.64} $\pm 0.38$ & \textbf{85.38}   $\pm 0.50$       \\ \hline
        
        EnergyOE       & 74.50 $\pm 0.16$ & \textbf{34.96} $\pm 0.89$ & \textbf{89.05} $\pm 0.26$ & 85.46 $\pm 0.38$          \\ 
        \textbf{EnergyOE+Ours}           & \textbf{74.93} $\pm 0.21$ & 35.12 $\pm 1.35$ & \textbf{89.05} $\pm 0.59$ & \textbf{85.47} $\pm 0.92$        \\ \hline
        
        OECC       & 69.36 $\pm 0.21$ & 36.98 $\pm 1.39$ & 87.51 $\pm 0.48$ & 82.90 $\pm 0.69$          \\ 
        
        \textbf{OECC+Ours}   & \textbf{72.23} $\pm 0.10$ & \textbf{33.76} $\pm 1.35$ & \textbf{89.51} $\pm 0.40$ & \textbf{86.41}   $\pm 0.61$       \\ \hline
       \end{tabular}
\label{MOOD_balanced_cifar100_std}
\end{subtable}
\label{MOOD_comparison_balanced_std}
\end{table}
\newpage

{\bf\noindent LT-CIFAR model on MOOD:}
\begin{table}[h!]
\caption{Comparison result on MOOD benchmark using ResNet18; Mean and std over 8 random runs are reported; MOOD benchmark detection average (over 11 datasets) performance (FPR95, AUROC, AUPR) and classification accuracy (ACC) (a): Result on LT-CIFAR10 (b): Result on LT-CIFAR100.}
\begin{subtable}{1.0\linewidth}
\centering
\scriptsize
\caption{}
\label{MOOD_imbalanced_cifar10_std}
\begin{tabular}{c|c|c|c|c}
\hline

Method         &ACC$\uparrow$     &FPR$\downarrow$     &      AUROC$\uparrow$   &     AUPR$\uparrow$         \\ \hline
MSP(ST)         & 73.28 $\pm 0.00$         & 57.72 $\pm 0.00$   & 75.96 $\pm 0.00$    & 69.93 $\pm 0.00$     \\

 Energy(ST)      & 73.28 $\pm 0.00$ & \textbf{43.91} $\pm 0.00$      & \textbf{83.88} $\pm 0.00$  & 78.40 $\pm 0.00$     \\
 ReAct(ST)~\cite{sun2021react}   & 73.28 $\pm 0.00$ & 43.95 $\pm 0.00$ & 83.88 $\pm 0.00$ & 78.40 $\pm 0.00$  \\
 kNN(ST)~\cite{sun2022out}   & 73.28 $\pm 0.00$ & 65.40 $\pm 0.00$ & 81.32 $\pm 0.00$ & \textbf{80.64} $\pm 0.00$  \\
  \hline
  OE(scratch)       & 72.85 $\pm 1.26$        & 35.11 $\pm 2.04$    & 88.07 $\pm 0.77$    & 82.79 $\pm 1.41$      \\ 
  
   EnergyOE(scratch) & 73.40 $\pm 0.52$        & \textbf{33.99} $\pm 1.33$   & 84.45 $\pm 0.86$      & 74.61 $\pm 1.86$ \\
   
   PASCL~\cite{wang2022partial}     & \textbf{77.08} $\pm 1.01$    &  35.88 $\pm 1.26$  & \textbf{88.90} $\pm 0.56$   &  \textbf{ 85.42} $\pm 1.14$     \\   
   
      OS~\cite{wei2022open}   & 77.04 $\pm 0.54$   &   35.21 $\pm 1.94$  & 88.32 $\pm 0.73$       & 82.19 $\pm 1.60$   \\ \hline   

    OE      & 69.73 $\pm 0.12$ & 40.71 $\pm 0.20$ & 87.39 $\pm 0.07$ & 85.48 $\pm 0.08$    \\
    
        \textbf{OE+Ours}           & \textbf{77.66} $\pm 0.11$ & \textbf{30.46} $\pm 0.43$ & \textbf{91.14} $\pm 0.09$ & \textbf{89.88}  $\pm 0.11$        \\ \hline
        
        EnergyOE  & 74.77 $\pm 0.20$ & 31.22 $\pm 0.57$ & 90.99 $\pm 0.14$ & \textbf{89.87} $\pm 0.17$ \\ 
        
        \textbf{EnergyOE+Ours}           & \textbf{78.18} $\pm 0.13$ & \textbf{28.67} $\pm 0.71$ & \textbf{91.29} $\pm 0.19$ & 89.73 $\pm 0.25$         \\ \hline
        
        OECC        & 62.01 $\pm 0.26$ & 34.78 $\pm 0.23$ & 89.48 $\pm 0.08$ & \textbf{87.94} $\pm 0.17$  \\ 
        
        \textbf{OECC+Ours}     & \textbf{75.23} $\pm 0.16$ & \textbf{30.06} $\pm 0.89$ & \textbf{90.53} $\pm 0.25$ & 87.73  $\pm 0.51$   \\\hline 
        
        BEnergyOE & 76.30 $\pm 0.24$ & 27.59 $\pm 0.59$ & 91.91 $\pm 0.15$ & \textbf{90.46} $\pm 0.18$  \\ 
        
        \textbf{BEnergyOE   +Ours}   & \textbf{78.13} $\pm 0.18$ & \textbf{25.99} $\pm 0.48$ & \textbf{92.09} $\pm 0.13$ & \textbf{90.46} $\pm 0.19$         \\         
        \hline
       \end{tabular}
\end{subtable}
\begin{subtable}{1.0\linewidth}
\centering
\scriptsize
\caption{}
\begin{tabular}{c|c|c|c|c}
\hline
Method         &ACC$\uparrow$     &FPR$\downarrow$     &      AUROC$\uparrow$   &     AUPR$\uparrow$         \\ \hline
MSP(ST)         & \textbf{40.22} $\pm 0.00$         & 78.68 $\pm 0.00$     & 64.16 $\pm 0.00$       & 58.89 $\pm 0.00$         \\

 Energy(ST)      & \textbf{40.22} $\pm 0.00$   & 69.56 $\pm 0.00$  & 69.66 $\pm 0.00$   & 63.17 $\pm 0.00$   \\
 ReAct(ST)~\cite{sun2021react}   & 39.02 $\pm 0.00$ & \textbf{69.01} $\pm 0.00$ & 70.32 $\pm 0.00$ & 63.73 $\pm 0.00$  \\
 kNN(ST)~\cite{sun2022out}   & \textbf{40.22} $\pm 0.00$ & 82.82 $\pm 0.00$ & \textbf{72.14} $\pm 0.00$ & \textbf{72.41} $\pm 0.00$  \\
  \hline
  OE(scratch)       & 41.31 $\pm 0.66$         & 60.36 $\pm 3.73$    & 76.95 $\pm 2.03$     & 71.28 $\pm 2.46$         \\ 
  
   EnergyOE(scratch) & 40.78 $\pm 0.33$        &  \textbf{52.20} $\pm 2.07$ & \textbf{79.41} $\pm 1.16$   & \textbf{72.96} $\pm 1.59$ \\
   
   PASCL~\cite{wang2022partial}      & \textbf{43.10}  $\pm 0.47$      &    65.32 $\pm 1.03$     &   74.25  $\pm 0.87$     & 68.35 $\pm 1.02$    \\   
   
      OS~\cite{wei2022open}   & 39.96 $\pm 0.60$   & 58.56 $\pm 1.86$   &   77.25 $\pm 1.49$  &      70.72 $\pm 2.06$  \\ \hline   

    OE     & 38.93 $\pm 0.07$ & 66.27 $\pm 0.16$ & 74.36 $\pm 0.11$ & 69.28 $\pm 0.12$          \\
    
        \textbf{OE+Ours}    & \textbf{39.73} $\pm 0.09$ & \textbf{51.04} $\pm 0.51$ & \textbf{81.43} $\pm 0.22$ & \textbf{76.56} $\pm 0.24$    \\ \hline
        
        EnergyOE        & 40.54 $\pm 0.08$ & 50.05 $\pm 0.30$ & 81.82 $\pm 0.18$ & \textbf{77.17} $\pm 0.26$          \\ 
        
        \textbf{EnergyOE+Ours}           & \textbf{40.76} $\pm 0.13$ & \textbf{47.81} $\pm 0.42$ & \textbf{82.31} $\pm 0.23$ & 76.49 $\pm 0.35$         \\ \hline
        
        OECC      & 31.08 $\pm 0.11$ & 50.75 $\pm 0.20$ & 81.79 $\pm 0.12$ & \textbf{77.90} $\pm 0.17$  \\ 
        
        \textbf{OECC+Ours}      & \textbf{39.77} $\pm 0.13$ & \textbf{47.28} $\pm 0.40$ & \textbf{82.52} $\pm 0.12$ & 77.68 $\pm 0.18$   \\\hline 
        
        BEnergyOE   & 40.78 $\pm 0.21$ & 45.37 $\pm 0.42$ & 83.64 $\pm 0.21$ & 79.21 $\pm 0.28$    \\ 
        
        \textbf{BEnergyOE   +Ours}  & \textbf{43.45} $\pm 0.14$ & \textbf{43.06} $\pm 0.44$ & \textbf{84.78} $\pm 0.17$ & \textbf{80.68} $\pm 0.27$         \\         
        \hline
       \end{tabular}
\label{MOOD_imbalanced_cifar100_std}
\end{subtable}
\label{MOOD_comparison_imbalanced_std}
\end{table}
\newpage

\subsection{Detailed Result on SYN with Standard Deviation}
\label{detalied_result_SYN}

In comparison to MOOD and SC-OOD benchmarks, the SYN benchmark shows relatively less stability in terms of performance improvement. However, when comparing OE with our algorithm, we still observe performance enhancement in 11 out of 12 OOD detection performance metrics, including AUROC, FPR, and AUPR, across the four in-distribution datasets within the SYN benchmark.

{\bf\noindent CIFAR model on SYN:}
\begin{table}[h!]
\caption{Comparison result on SYN benchmark using ResNet18; Mean and std over eight random runs are reported; SYN benchmark detection average (over 3 datasets) performance (FPR95, AUROC, AUPR) and classification accuracy (ACC) (a): Result on CIFAR10 (b): Result on CIFAR100.}
\begin{subtable}{1.0\linewidth}
\centering
\scriptsize
\caption{}
\label{syn_balanced_cifar10_std}
\begin{tabular}{c|c|c|c|c}
\hline

Method         &ACC$\uparrow$     &FPR$\downarrow$     &      AUROC$\uparrow$   &     AUPR$\uparrow$         \\ \hline
MSP(ST)         & 93.69  $\pm 0.00$        & 20.07 $\pm 0.00$     & 90.73 $\pm 0.00$     & 83.14 $\pm 0.00$      \\

 Energy(ST)      & 93.69 $\pm 0.00$         & 14.36 $\pm 0.00$  & 93.38 $\pm 0.00$ & 86.37 $\pm 0.00$    \\
 ReAct(ST)~\cite{sun2021react}   & \textbf{93.70} $\pm 0.00$ & \textbf{13.74} $\pm 0.00$ & 93.69 $\pm 0.00$ & 86.82 $\pm 0.00$  \\
 kNN(ST)~\cite{sun2022out}   & 93.69 $\pm 0.00$ & 32.09 $\pm 0.00$ & \textbf{96.80} $\pm 0.00$ & \textbf{95.14} $\pm 0.00$  \\ \hline
 
  OE(scratch)       & 90.12 $\pm 0.69$         & 2.39 $\pm 2.35$    & \textbf{99.40} $\pm 0.50$    & \textbf{99.14} $\pm 0.56$       \\ 
  
   EnergyOE(scratch) & 90.50 $\pm 0.35$     & \textbf{1.82} $\pm 1.33$  & 99.26 $\pm 0.71$ & 98.44 $\pm 1.61$ \\
   
   WOOD(10\%)     & 93.18 $\pm 0.00$    &  4.05 $\pm 0.00$   &  98.69 $\pm 0.00$ &   97.20 $\pm 0.00$ \\  
   
   WOOD(100\%)     & 93.2 $\pm 0.00$   &  3.80 $\pm 0.00$ & 98.72 $\pm 0.00$ &   97.16 $\pm 0.00$  \\ 
   
      NTOM(10\%)   & 84.37 $\pm 2.08$  &  27.65 $\pm 6.90$  &  87.18 $\pm 5.03$     & 80.12 $\pm 6.61$   \\
      
      NTOM(100\%)   & \textbf{94.39} $\pm 0.24$   & 10.33 $\pm 1.75$ & 95.14 $\pm 1.00$  &  90.80 $\pm 2.11$  \\\hline   

    OE   & 93.12 $\pm 0.05$ & 3.97 $\pm 0.13$ & 98.77 $\pm 0.05$ & \textbf{97.63} $\pm 0.13$   \\ 
    
        \textbf{OE+Ours}     & \textbf{93.38} $\pm 0.10$ & \textbf{2.18} $\pm 0.29$ &\textbf{98.80} $\pm 0.19$ & 96.24 $\pm 0.54$ \\ \hline
        
        EnergyOE       & 93.33 $\pm 0.09$ & 1.67 $\pm 0.24$ &\textbf{ 99.28 } $\pm 0.10$ & \textbf{97.90} $\pm 0.31$    \\ 
        
        \textbf{EnergyOE+Ours}           & \textbf{93.64} $\pm 0.06$ & \textbf{1.52} $\pm 0.20$ & 99.24 $\pm 0.10$ & 97.63  $\pm 0.31$        \\ \hline
        
        OECC    & 91.81 $\pm 0.24$ & \textbf{1.49} $\pm 0.26$ & \textbf{99.31} $\pm 0.20$ & \textbf{97.87} $\pm 0.86$   \\ 
        
        \textbf{OECC+Ours}    & \textbf{92.18} $\pm 0.28$ & 3.86 $\pm 0.35$ & 97.55 $\pm 0.38$ & 93.11 $\pm 1.29$  \\ \hline
       \end{tabular}
\end{subtable}
\begin{subtable}{1.0\linewidth}
\centering
\scriptsize
\caption{}
\begin{tabular}{c|c|c|c|c}
\hline

Method         &ACC$\uparrow$     &FPR$\downarrow$     &      AUROC$\uparrow$   &     AUPR$\uparrow$         \\ \hline
MSP(ST)         & \textbf{75.70}  $\pm 0.00$        & 39.91 $\pm 0.00$   & 78.27 $\pm 0.00$  & 67.49 $\pm 0.00$  \\

 Energy(ST)      & \textbf{75.70} $\pm 0.00$ & 49.52 $\pm 0.00$  & 61.58 $\pm 0.00$  & 56.77 $\pm 0.00$ \\
 ReAct(ST)~\cite{sun2021react}   & 74.79 $\pm 0.00$ & \textbf{25.72} $\pm 0.00$ & 83.11 $\pm 0.00$ & 70.31 $\pm 0.00$  \\
 kNN(ST)~\cite{sun2022out}   & \textbf{75.70}  $\pm 0.00$ & 46.24 $\pm 0.00$ & \textbf{87.12} $\pm 0.00$ & \textbf{85.99} $\pm 0.00$  \\ \hline
 
  OE(scratch)       & 66.47 $\pm 0.45$  & 13.67 $\pm 13.10$  & 93.96 $\pm 7.78$ & \textbf{93.09} $\pm 7.73$     \\ 
  
   EnergyOE(scratch) & 65.92 $\pm 0.77$          & \textbf{10.99} $\pm 8.64$   & \textbf{95.24} $\pm 5.20$    & 92.93 $\pm 6.33$ \\
   
   WOOD(10\%)     & 75.8 $\pm 0.00$     &  35.13 $\pm 0.00$ & 73.93 $\pm 0.00$  &  69.35 $\pm 0.00$ \\  
   
   WOOD(100\%)     & \textbf{75.93} $\pm 0.00$   &   35.16 $\pm 0.00$ &   73.96 $\pm 0.00$     & 69.41 $\pm 0.00$   \\ 
   
      NTOM(10\%)   & 49.81 $\pm 4.43$  & 74.72 $\pm 9.72$  & 43.00 $\pm 10.23$    &      49.42 $\pm 4.98$ \\
      
      NTOM(100\%)   &  72.41 $\pm 0.95$ & 37.15 $\pm 6.23$  & 80.21 $\pm 7.08$    & 72.86 $\pm 7.07$  \\\hline  

    OE   & \textbf{73.86} $\pm 0.10$  & 13.03 $\pm 0.67$ & 91.94 $\pm 0.52$ & 86.48 $\pm 0.53$    \\ 
    
        \textbf{OE+Ours}    & 73.13 $\pm 0.19$ & \textbf{3.78} $\pm 0.49$ & \textbf{98.55} $\pm 0.19$ & \textbf{96.48} $\pm 0.43$    \\ \hline
        
        EnergyOE   & 74.50 $\pm 0.16$ & 5.34 $\pm 1.44$ & 98.02 $\pm 0.53$ & 95.82 $\pm 1.02$  \\
        
        \textbf{EnergyOE+Ours}           & \textbf{74.93} $\pm 0.21$ & \textbf{2.85} $\pm 0.70$ & \textbf{99.17} $\pm 0.25$ & \textbf{98.37} $\pm 0.64$        \\ \hline
        
        OECC     & 69.36 $\pm 0.21$ & 6.71 $\pm 1.10$ & 96.67 $\pm 0.62$ & 92.23 $\pm 1.53$         \\ 
        \textbf{OECC+Ours}     & \textbf{72.23} $\pm 0.10$ & \textbf{5.94} $\pm 0.85$ & \textbf{96.98} $\pm 0.55$ & \textbf{93.08} $\pm 1.48$    \\ \hline
       \end{tabular}
\label{balanced_cifar100_std}
\end{subtable}
\end{table}
\newpage

{\bf\noindent LT-CIFAR model on SYN:}
\begin{table}[h!]
\caption{Comparison result on SYN benchmark using ResNet18; Mean and std over eight random runs are reported; SYN benchmark detection average (over 3 datasets) performance (FPR95, AUROC, AUPR) and classification accuracy (ACC) (a): Result on LT-CIFAR10 (b): Result on LT-CIFAR100.}
\begin{subtable}{1.0\linewidth}
\centering
\scriptsize
\caption{}
\label{syn_imbalanced_cifar10_std}
\begin{tabular}{c|c|c|c|c}
\hline

Method         &ACC$\uparrow$     &FPR$\downarrow$     &      AUROC$\uparrow$   &     AUPR$\uparrow$         \\ \hline
MSP(ST)         & 73.28  $\pm 0.00$        & 48.73 $\pm 0.00$  & \textbf{76.76} $\pm 0.00$  & \textbf{70.40} $\pm 0.00$    \\

 Energy(ST)      & 73.28 $\pm 0.00$ & \textbf{45.86} $\pm 0.00$   & 74.76 $\pm 0.00$ & 63.16 $\pm 0.00$  \\
 ReAct(ST)~\cite{sun2021react}  & 73.28 $\pm 0.00$ & 45.87 $\pm 0.00$ & 74.75 $\pm 0.00$ & 63.16 $\pm 0.00$  \\
 kNN(ST)~\cite{sun2022out}   & 73.28 $\pm 0.00$ & 84.89 $\pm 0.00$ & 76.34 $\pm 0.00$ & 65.11 $\pm 0.00$  \\ \hline
 
  OE(scratch)       & 72.85 $\pm 1.26$    & 18.67 $\pm 4.69$   & 91.58 $\pm 1.60$ & 83.77 $\pm 2.68$    \\ 
  
   EnergyOE(scratch) & 73.40 $\pm 0.52$   & 13.60 $\pm 3.82$   & 93.90 $\pm 2.41$   & 88.83 $\pm 4.79$ \\
   
   PASCL~\cite{wang2022partial}     & \textbf{77.08} $\pm 1.01$   &  8.49 $\pm 2.32$  & \textbf{96.53} $\pm 1.08$  &      \textbf{93.33} $\pm 2.23$    \\   
   
      OS~\cite{wei2022open}   & 77.04 $\pm 0.54$  &  \textbf{8.28} $\pm 1.43$  & 94.54 $\pm 1.15$  & 86.43 $\pm 2.62$  \\ \hline   

    OE        & 69.73 $\pm 0.12$ & 30.27 $\pm 0.10$ & 80.03 $\pm 0.34$ & 77.69 $\pm 0.11$    \\
    
        \textbf{OE+Ours}           & \textbf{77.66} $\pm 0.11$ & \textbf{20.90} $\pm 0.39$ & \textbf{92.32} $\pm 0.27$ & \textbf{87.79} $\pm 0.49$    \\ \hline
        
        EnergyOE    & 74.77 $\pm 0.20$ & 9.67 $\pm 0.85$ & 96.60 $\pm 0.38$ & 94.26 $\pm 0.67$ \\ 
        
        \textbf{EnergyOE+Ours}    & \textbf{78.18} $\pm 0.13$ &\textbf{ 7.79 } $\pm 0.78$ & \textbf{97.35} $\pm 0.24$ &\textbf{ 95.01} $\pm 0.46$    \\ \hline
        
        OECC         & 62.01 $\pm 0.26$ & 21.69 $\pm 0.38$ & 90.80 $\pm 0.34$ & 87.84 $\pm 0.64$  \\ 
        
        \textbf{OECC+Ours}     & \textbf{75.23} $\pm 0.16$ & \textbf{15.49} $\pm 1.56$ & \textbf{94.96} $\pm 0.33$ & \textbf{91.70}  $\pm 0.44$   \\\hline 
        
        BEnergyOE& 76.30 $\pm 0.24$ & 7.52 $\pm 0.79$ & 97.51 $\pm 0.32$ & 95.23 $\pm 0.64$  \\ 
        
        \textbf{BEnergyOE   +Ours}   & \textbf{78.13} $\pm 0.18$ & \textbf{6.85} $\pm 0.97$ & \textbf{97.81} $\pm 0.27$ & \textbf{95.58} $\pm 0.40$         \\         
        \hline
       \end{tabular}
\end{subtable}
\begin{subtable}{1.0\linewidth}
\centering
\scriptsize
\caption{}
\begin{tabular}{c|c|c|c|c}
\hline
Method         &ACC$\uparrow$     &FPR$\downarrow$     &      AUROC$\uparrow$   &     AUPR$\uparrow$         \\ \hline
MSP(ST)         & \textbf{40.22} $\pm 0.00$         & 81.44 $\pm 0.00$   & 39.46 $\pm 0.00$   & 45.89 $\pm 0.00$    \\

 Energy(ST)      & \textbf{40.22} $\pm 0.00$   & 74.98 $\pm 0.00$ & 40.91 $\pm 0.00$  & 45.33 $\pm 0.00$  \\
 ReAct(ST)~\cite{sun2021react}   & 39.02 $\pm 0.00$ & 62.00 $\pm 0.00$ & 55.96 $\pm 0.00$ & 50.39 $\pm 0.00$  \\
 kNN(ST)~\cite{sun2022out}   & \textbf{40.22} $\pm 0.00$ & \textbf{14.20} $\pm 0.00$ & \textbf{96.83} $\pm 0.00$ & \textbf{92.96} $\pm 0.00$  \\ \hline
 
  OE(scratch)       & 41.31  $\pm 0.66$     & 22.66 $\pm 5.52$    & 85.76 $\pm 2.27$   & 72.73 $\pm 2.48$       \\ 
  
   EnergyOE(scratch) & 40.78 $\pm 0.33$     &  18.54 $\pm 7.22$  & 90.10 $\pm 6.34$   & 86.07 $\pm 8.57$ \\
   
   PASCL~\cite{wang2022partial}     & \textbf{43.10} $\pm 0.47$  & 30.07 $\pm 6.97$  & 82.13 $\pm 4.60$  & 69.88 $\pm 4.90$  \\   
   
      OS~\cite{wei2022open}   & 39.96 $\pm 0.60$   & \textbf{16.12} $\pm 4.04$   & \textbf{93.54} $\pm 2.27$   &  \textbf{88.07} $\pm 4.42$   \\ \hline   

    OE        & 38.93 $\pm 0.07$ & 72.83 $\pm 1.02$ & 58.62 $\pm 1.22$ & 52.99 $\pm 0.66$    \\
    
        \textbf{OE+Ours}   & \textbf{39.73} $\pm 0.09$ & \textbf{28.87} $\pm 2.03$ & \textbf{86.40} $\pm 1.39$ & \textbf{77.65} $\pm 1.58$   \\ \hline
        
        EnergyOE & 40.54 $\pm 0.08$ & 34.89 $\pm 3.20$ & 84.98 $\pm 2.36$ & 76.71 $\pm 3.04$   \\ 
        
        \textbf{EnergyOE+Ours}    & \textbf{40.76} $\pm 0.13$ & \textbf{10.14} $\pm 0.89$ & \textbf{97.11} $\pm 0.27$ & \textbf{95.81} $\pm 0.41$  \\ \hline
        
        OECC      & 31.08 $\pm 0.11$ & \textbf{5.91} $\pm 0.15$ &\textbf{ 96.87} $\pm 0.05$ & \textbf{91.74 } $\pm 0.13$  \\ 
        
        \textbf{OECC+Ours}     & \textbf{39.77} $\pm 0.13$ & 10.81 $\pm 0.53$ & 94.56 $\pm 0.28$ & 88.36 $\pm 0.45$   \\\hline 
        
        BEnergyOE  & 40.78 $\pm 0.21$ & 15.84 $\pm 3.15$ & 93.82 $\pm 1.81$ & 88.40 $\pm 2.81$    \\ 
        
        \textbf{BEnergyOE   +Ours}  & \textbf{43.45} $\pm 0.14$ & \textbf{3.68} $\pm 0.26$ & \textbf{98.21} $\pm 0.15$ &\textbf{ 94.90 } $\pm 0.38$        \\         
        \hline
       \end{tabular}
\label{syn_imbalanced_cifar100_std}
\end{subtable}
\end{table}

\newpage

\section*{ \large Network Analysis}
\subsection{Generality of our method on other Network}

\begin{table}[h!]
\caption{Comparison result on SC-OOD benchmark using WideResNet40-2; Mean and std over eight random runs are reported; SC-OOD benchmark detection average (over 6 datasets) performance (FPR95, AUROC, AUPR) and classification accuracy (ACC) (a): Result on LT-CIFAR10 (b): Result on LT-CIFAR100.}
\begin{subtable}{1.0\linewidth}
\centering
\scriptsize
\caption{}
\label{syn_imbalanced_cifar10_std}
\begin{tabular}{c|c|c|c|c}
\hline

Method         &ACC$\uparrow$     &FPR$\downarrow$     &      AUROC$\uparrow$   &     AUPR$\uparrow$         \\ \hline
MSP(ST)     & 72.32 $\pm 0.00$  & 63.96 $\pm 0.00$  & 74.60 $\pm 0.00$  & 72.62 $\pm 0.00$   \\

 Energy(ST)  & 72.32 $\pm 0.00$ & 58.44 $\pm 0.00$ & 80.23 $\pm 0.00$ & 77.67 $\pm 0.00$  \\ \hline

 OE & 69.21 $\pm 0.13$ & 54.49 $\pm 0.27$ & 83.98 $\pm 0.09$ & 83.75 $\pm 0.06$  \\
 \textbf{OE+Ours} & \textbf{76.64} $\pm 0.24$ & \textbf{40.12} $\pm 0.27$ & \textbf{89.91} $\pm 0.04$ & \textbf{89.12} $\pm 0.03$  \\ \hline

 OECC & 50.69 $\pm 0.50$ & 55.02 $\pm 0.37$ & 82.77 $\pm 0.16$ & 81.69 $\pm 0.19$  \\
 \textbf{OECC+Ours} & \textbf{73.86} $\pm 0.44$ & \textbf{34.57} $\pm 0.40$ & \textbf{90.10} $\pm 0.25$ & \textbf{87.89} $\pm 0.52$  \\ \hline

 EnergyOE & 74.82 $\pm 0.33$ & 34.12 $\pm 0.27$ & 91.41 $\pm 0.07$ & 90.96 $\pm 0.07$  \\
 \textbf{EnergyOE+Ours} & \textbf{77.33} $\pm 0.29$ & \textbf{32.69} $\pm 0.24$ & \textbf{91.72} $\pm 0.06$ & \textbf{91.36} $\pm 0.12$  \\ \hline

 BEnergyOE & \textbf{75.69} $\pm 0.23$ & 31.23 $\pm 0.40$ & 91.78 $\pm 0.06$  & 90.45 $\pm 0.17$  \\
 \textbf{BEnergyOE+Ours} & 75.64 $\pm 0.31$ & \textbf{30.86} $\pm 0.24$ & \textbf{91.90} $\pm 0.07$ & \textbf{90.70} $\pm 0.10$  \\ \hline

       \end{tabular}
\end{subtable}
\begin{subtable}{1.0\linewidth}
\centering
\scriptsize
\caption{}
\begin{tabular}{c|c|c|c|c}
\hline
Method         &ACC$\uparrow$     &FPR$\downarrow$     &      AUROC$\uparrow$   &     AUPR$\uparrow$         \\ \hline
MSP(ST)     & 40.96 $\pm 0.00$ & 83.52 $\pm 0.00$ & 60.24 $\pm 0.00$  &  57.21 $\pm 0.00$  \\

 Energy(ST)  & 40.96 $\pm 0.00$ & 81.95 $\pm 0.00$ & 63.30 $\pm 0.00$ & 59.44 $\pm 0.00$  \\ \hline

 OE & 41.53 $\pm 0.06$ & 76.83 $\pm 0.11$ & 68.61 $\pm 0.05$ & 65.16 $\pm 0.06$  \\
 \textbf{OE+Ours} & \textbf{41.89} $\pm 0.10$ & \textbf{74.53} $\pm 0.11$ & \textbf{72.33} $\pm 0.04$ & \textbf{69.97} $\pm 0.04$  \\ \hline

 OECC & 30.54 $\pm 0.09$ & 69.84 $\pm 0.11$ & 72.42 $\pm 0.06$ & 69.02 $\pm 0.06$  \\
 \textbf{OECC+Ours} & \textbf{38.53} $\pm 0.12$ & \textbf{63.70} $\pm 0.20$ & \textbf{76.13} $\pm 0.10$ & \textbf{72.08} $\pm 0.10$  \\ \hline

 EnergyOE & 39.84 $\pm 0.08$ & 65.77 $\pm 0.33$ & 76.33 $\pm 0.16$ & 72.63 $\pm 0.15$  \\
 \textbf{EnergyOE+Ours} & \textbf{40.54} $\pm 0.24$ & \textbf{60.64} $\pm 0.25$ & \textbf{77.75} $\pm 0.10$ & \textbf{73.89} $\pm 0.10$  \\ \hline

 BEnergyOE & 39.36 $\pm 0.15$ & 62.95 $\pm 0.26$ & 77.21 $\pm 0.15$ & 72.87 $\pm 0.17$  \\
 \textbf{BEnergyOE+Ours} & \textbf{41.53} $\pm 0.11$ & \textbf{62.54} $\pm 0.32$ & \textbf{77.69} $\pm 0.10$ & \textbf{73.67} $\pm 0.09$  \\ \hline
       \end{tabular}
\label{syn_imbalanced_cifar100_std}
\end{subtable}
\label{wideresnet_total}
\end{table}

To demonstrate the generality of our method across different network architectures, we validate our results using WideResNet (WRN-40-2)~\cite{zagoruyko2016wide} instead of ResNet18. Table~\ref{wideresnet_total} illustrates the overall impact of our approach. Particularly in terms of OE and OECC, our method yields significant enhancements in classification accuracy and OOD detection performance. Notably, our method consistently enhances the OOD detection performance across all tuning-based methods on WideResNet.

\end{document}